\documentclass[lettersize,journal]{IEEEtran}

\usepackage{tabularx,booktabs}
\usepackage{float}
\usepackage{amsthm}
\usepackage{nccmath}
\usepackage{amssymb}
\usepackage{mathtools} 
\usepackage{bm}
\usepackage{caption}

\usepackage{amsmath,amsfonts}
\usepackage{array}
\usepackage[caption=false,font=normalsize,labelfont=sf,textfont=sf]{subfig}
\usepackage{textcomp}
\usepackage{multirow}
\usepackage{stfloats}
\usepackage{url}
\usepackage{verbatim}
\usepackage{graphicx}

\usepackage[ruled,vlined]{algorithm2e}
\makeatletter
\newcommand{\removelatexerror}{\let\@latex@error\@gobble}
\makeatother

\graphicspath{ {./images/} }
\usepackage{afterpage}

\begin{document}

\title{GeoFusion: An Uncertain Machine Learning Framework for Sensor Fusion}

\author{Basant Ahmed Mounir Abdelhamid and Amr T. Abdel-Hamid, \IEEEmembership{Senior Member, IEEE}
        % <-this % stops a space
\thanks{This is the authors' version of an article published in \emph{IEEE Access}, vol.~14, 2026, DOI: \protect\url{https://doi.org/10.1109/ACCESS.2026.3750664}. \copyright~2026 The Authors. This work is licensed under a Creative Commons Attribution 4.0 License. For more information, see \protect\url{https://creativecommons.org/licenses/by/4.0/}.}
\thanks{The first author is with the Faculty of Media Engineering and Technology, German University in Cairo (GUC), Cairo, Egypt (e-mail: basant.mounir@gmail.com). The second author is with King Abdullah University of Science and Technology (KAUST), Thuwal, Saudi Arabia.(e-mail: amr.abdelhamid@kaust.edu.sa)}% <-this % stops a space
}

% The paper headers
\markboth{Published in IEEE Access, vol.~14, 2026, DOI: 10.1109/ACCESS.2026.3750664}%
{Abdelhamid and Abdel-Hamid: GeoFusion: An Uncertain Machine Learning Framework for Sensor Fusion}

\maketitle

\begin{abstract}
High positioning precision is a critical requirement in many Global Positioning System (GPS) applications, yet achieving it with affordable smartphone sensors remains an open challenge. Existing approaches treat each GPS observation as an absolute point, discarding the uncertainty inherent in the measurement conditions and leaving the cooperative potential of multiple phones observing the same location untapped. In this work, we propose GeoFusion, a three-stage uncertain machine learning framework for cooperative sensor fusion. In the first stage, each GPS observation is modeled as an uncertain object using one of two proposed models: the \textit{volcano function}, grounded in satellite geometry, and the empirically derived \textit{mountain function}, built from the statistical structure of real smartphone data. In the second stage, uncertain objects are clustered by location event using partitional and graph-based algorithms. In the third stage, the cluster output is refined using filter-based and learning-based estimators. Experiments on the Google Smartphone Decimeter Challenge dataset show that uncertainty modeling is a primary driver of precision gains: the mountain function significantly outperforms the geometry-only volcano function, and uncertainty-aware UK-medoids significantly outperforms its certain counterpart, k-medoids. A post-clustering deep neural network further refines the cluster estimates, and the recommended configuration achieves a mean absolute error of 2.90 $\pm$ 0.15 meters on held-out test drives, comparable to and on average below the best reported single-phone precision of approximately 3 meters, demonstrating the potential of cooperative uncertainty-aware fusion for affordable GPS sensors.
\end{abstract}


\section{Introduction}
\label{sec:introduction}

\IEEEPARstart{O}{ver} the years, critical applications requiring Global Positioning System (GPS) high precision have emerged, including, but not limited to, precision agriculture \cite{agric}, fleet management \cite{civil}, road construction \cite{civil}, disaster management \cite{disaster}, underwater mapping \cite{survey}, and infrastructure surveying \cite{survey}. Efforts to meet these precision needs were traditionally driven by innovation in the logical and physical design of the sensors; areas of focus included antenna technology, filtering mechanisms, and positioning algorithms \cite{phylog}. While these efforts led to the development of GPS sensors delivering readings with high precision, such sensors are the most expensive on the market, easily exceeding the budget of many projects. Furthermore, if the system is already implemented and the precision requirements of an application change, physical intervention would be required, which may not be feasible.

To achieve higher GPS precision while preserving the underlying GPS sensor technology, researchers devised sensor fusion schemes. Sensor fusion consults data from multiple available sensors instead of just one for better positioning. One popular approach is to fuse data from different types of sensors of the same system instance, or \textit{observer}. In \cite{inertiagps}, GPS and Inertial Navigation System (INS) data from an observer are fused to more accurately position the observer. This approach assumes that more than one type of sensor is readily available in each observer, which may not be the case. Even if such sensors are present or their integration is feasible, the observers must possess either the compute resources to execute the fusion algorithm or the bandwidth to communicate all the data to another location for fusion. In some contexts, edge computing may not be feasible, or the bandwidth may not be large enough to transmit all the data due to hostile conditions. Another approach is to fuse the same sensor data from multiple observers. Using vehicular ad hoc network (VANET) technology, researchers in \cite{lica} improved GPS positioning precision of individual vehicles by fusing data from nearby vehicles. A key limitation of this approach is that precision cannot be improved if there are not enough nearby vehicles. In this scenario, a vehicle cannot consult data from vehicles that crossed the same location in the past.

With the disruptive advent of machine learning, a new research direction for improving GPS precision emerged. Machine learning can help mitigate errors by leveraging data, making higher precision accessible even with lower technological sensor capabilities. For example, using signal reception data, classifiers were trained to mitigate pseudorange noise in \cite{multipath1}, \cite{multipath2}, and \cite{multipath3}. Machine learning was also used to detect GPS satellite oscillator anomalies in \cite{oscilerror}. The detection of ionospheric scintillation events that may impact GPS precision was implemented in \cite{ionospherror}. More recently, deep learning approaches have demonstrated that a GPS baseline position can be refined by a learned corrector mapping receiver-observable quality features to a position offset, achieving promising results on smartphone data \cite{siemuri2021}, \cite{kanhere2022improving}, \cite{pcdeepnet2025}. The best single-phone precision reported on real smartphone data is approximately 3 meters \cite{fu2020}, which serves as the baseline that cooperative fusion approaches such as GeoFusion aim to exceed. However, the data required to train the models may not occur naturally in the context of the application; this can complicate data acquisition and retraining efforts when drift occurs. 

The aforementioned approaches share three unresolved limitations. First, multi-sensor and VANET-based fusion methods require either multiple sensor types or contemporaneous nearby observers, and cannot exploit historical observations of the same location. Second, machine learning correctors operate on a single phone at a single epoch, leaving the cooperative potential of multiple phones observing the same location event untapped. Third, and most fundamentally, all existing approaches treat each GPS observation as a precise point, discarding the uncertainty inherent in the measurement conditions. While uncertain clustering algorithms exist in the data mining literature \cite{ukmeans}, \cite{ukmedoids}, no prior work has combined uncertainty representations of GPS observations with uncertain clustering to conduct cooperative sensor fusion for higher positioning precision. Recent graph-based clustering methods have advanced this direction further by learning structured doubly stochastic affinity matrices that jointly encode cluster membership and graph topology \cite{SDSGC, patcog2026,  sigpro110144, neucom130571}, yet these methods remain confined to the data mining domain and have not been applied to GPS sensor fusion. GeoFusion is designed to fill these gaps.

\begin{figure}[!t]
\centering
\includegraphics[width=8.4cm]{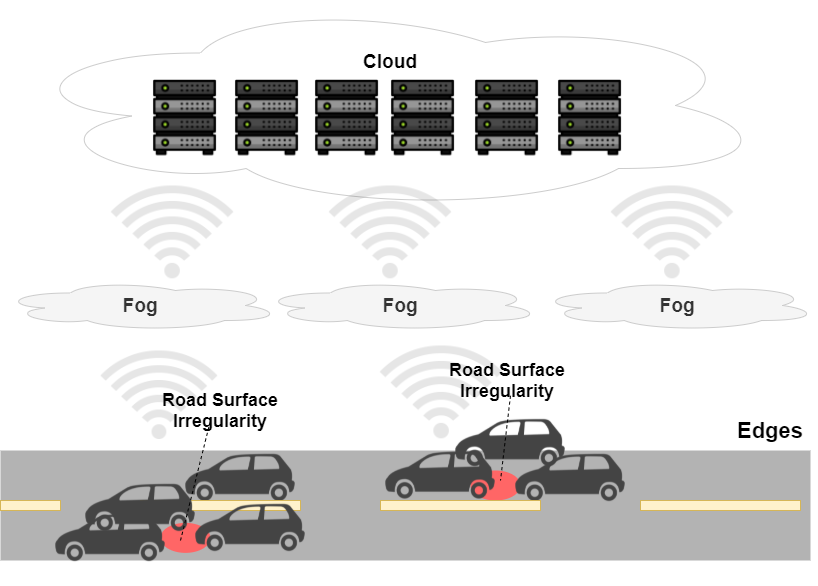}
\caption{Instead of relying on expensive scanning vans with high-precision GPS sensors, this architecture collects GPS data from vehicles during their regular commutes and processes it on the cloud. GeoFusion clusters the observations by location event, identifying the positions of road damage from cooperative smartphone GPS readings. This approach eliminates the need for intensive scheduling and road closures while achieving higher positioning precision.}
\label{img:geo-application}
\end{figure}

\IEEEpubidadjcol
In this work, we propose GeoFusion, a framework that combines techniques from machine learning, uncertainty representation, and sensor fusion to address the aforementioned limitations. GeoFusion is designed for any GPS-equipped device, with smartphones serving as the primary instantiation in this work. GeoFusion is designed to economically bolster GPS precision in accordance with changing precision requirements and without significantly disrupting or modifying system operations. Furthermore, only GPS readings are required; GeoFusion makes no other assumptions about the types of sensors present. Because GeoFusion only requires GPS readings, it may be well-suited for projects with limited communication bandwidth or edge computing capacity. In addition, GeoFusion uses the reported GPS data, which are naturally occurring in the context of systems with positioning modules; no special data acquisition efforts are required to continually improve the results. Historical data are also used to augment the estimation process with past conditions. GeoFusion has a mechanism to represent uncertainty in observations, capturing more information about the conditions in which the observation was reported. Fig.~\ref{img:geo-application} shows the GeoFusion framework applied in road surveying.

The underlying assumption of GeoFusion is that a \textit{location event}, a phenomenon whose location we want to detect, is \textit{observable} more than once; in other words, more than one smartphone observation of the same location event must have been previously reported. In cases where a location event is observed only once, the cluster representative defaults to the single reported phone position and no fusion benefit is realized for that event. This observability condition is the theoretical basis for GeoFusion's precision gains: by fusing $n_C$ independent noisy observations of the same true location within a cluster, the expected positioning error reduces at a rate of $O(1/\sqrt{n_C})$ \cite{kay1993}, meaning that more smartphone observations of a location event directly translate to higher precision estimates. GeoFusion also assumes that data transmission from smartphones to the cloud is supported. To the best of our knowledge, no other work has explored combining uncertainty representations of observations and uncertain clustering to conduct sensor fusion for higher sensor precision.

The GeoFusion framework consists of three stages: pre-clustering uncertainty representation, clustering observations by location event, and post-clustering location event position estimation. In the first stage, each smartphone GPS observation is modeled as an uncertain object, a probability distribution over all plausible true locations, weighted by the factors that contribute most to positioning error. Two uncertain object models are proposed: the \textit{volcano function}, grounded in satellite geometry, and the \textit{mountain function}, derived empirically from the statistical structure of the Google Smartphone Decimeter Challenge dataset. In the second stage, uncertain objects are grouped by location event using partitional clustering algorithms, namely k-means \cite{kmeans}, k-medoids \cite{kmedoids}, and their uncertainty-aware counterparts UK-means \cite{ukmeans} and UK-medoids \cite{ukmedoids}, as well as the graph-based Structured Doubly Stochastic Graph-Based Clustering (SDSGC) algorithm \cite{SDSGC}. In the third stage, the observations within each cluster are fused into a single position estimate using classic filter-based estimators, namely the Kalman filter (KF) \cite{kalman1960new}, extended Kalman filter (EKF) \cite{kalman1961new}, and particle filter (PF) \cite{pf}, as well as a learning-based post-clustering deep neural network (DNN) \cite{siemuri2021}. By incorporating uncertainty and harnessing the power of collective decision-making, GeoFusion uses affordable smartphones with a GPS precision of 3 to 5 meters to estimate positions with higher precision than any individual phone, exceeding the best single-phone precision of 3 meters \cite{fu2020} with an average MAE of $2.90$\,m on real smartphone data.

We make the following contributions:
\begin{itemize}
\item \textbf{The GeoFusion framework} for cooperative GPS sensor fusion using uncertain object representations, designed to exceed single-phone positioning precision using naturally occurring GPS data and without requiring physical sensor upgrades.
\item \textbf{Two uncertain object models for smartphone GPS observations}: the \textit{volcano function}, grounded in satellite geometry via the cost function of \cite{pdf}, and the empirically derived \textit{mountain function}, built directly from the statistical structure of real smartphone data.
\item \textbf{A preprocessing pipeline and publicly released real-world GNSS dataset}, derived from the Google Smartphone Decimeter Challenge, available at \url{https://www.kaggle.com/code/basantmounir/geofusion-dataset}.
\item \textbf{A systematic benchmarking study} of partitional and graph-based uncertain clustering algorithms across a range of scales, with statistically validated findings.
\item \textbf{A post-clustering estimation framework} combining filter-based and learning-based estimators evaluated on real smartphone data.
\end{itemize}

The paper proceeds as follows. Section~\ref{sec:dataset} describes the dataset and preprocessing pipeline. In Section~\ref{sec:geo}, the three stages of GeoFusion are described in detail. Section~\ref{sec:experiments} describes the experimental design and evaluation criteria, and presents the results. Finally, Section~\ref{sec:conc} concludes this work.
\section{Dataset}
\label{sec:dataset}

\subsection{Raw Dataset}
The dataset used in this study comes from the Google Smartphone Decimeter Challenge \cite{fu2020}. Real location data were collected from 29 drives, each with 1 to 4 phones simultaneously, over approximately one year across Mountain View, Sunnyvale, San Jose, Redwood City, and San Francisco using 7 distinct phone models. This provides location data across varying environmental conditions including urban canyons, tree-lined suburban roads, and open highways. In addition to the reported phone locations, reference positions are measured using a NovAtel SPAN system. While reported phone locations carry errors typically between 1 and 20 meters, the NovAtel system capitalizes on dual-frequency reception, real-time kinematics, and carrier-phase measurements to achieve centimeter-level horizontal accuracy \cite{fu2020, everett2022}.

The raw dataset comprises 29 folders with one folder per drive. For each drive folder, there are one to four subfolders, one per phone used to collect reported locations in that drive. Each phone folder contains three files:
\begin{itemize}
    \item \texttt{ground\_truth.csv} provides the NovAtel reference position for each epoch, against which phone positioning errors are measured, as well as the vehicle state and dilution of precision (DOP) captured by the GNSS engine of the phone.
    \item \texttt{\{phone\_name\}\_GnssLog.txt} contains raw signal-level measurements from the phone's GNSS chipset, including per-signal carrier-to-noise ratio, elevation, azimuth, and satellite usage flags.
    \item \texttt{\{phone\_name\}\_derived.csv} contains per-signal measurements processed from the GnssLog. Each row is identified by the drive, phone, epoch timestamp, satellite ID, and signal frequency, and captures satellite properties (ECEF position, velocity, clock bias), receiver-side measurements (raw pseudorange and its uncertainty), and atmospheric corrections (ionospheric and tropospheric delays).
\end{itemize}
The reported phone positions for each epoch are contained in \texttt{baseline\_locations\_train.csv} in the root directory. We use the training split exclusively, as the reference locations in the ground truth file are withheld for the test set.

\subsection{Preprocessing}
The raw data files were preprocessed to produce a single \texttt{.csv} file, each row representing an epoch per phone per drive. The following are the steps of the data preprocessing pipeline from the aforementioned raw files.

\begin{enumerate}

\item \textbf{Extract reported and reference locations.} From \texttt{baseline\_locations\_train.csv}, extract the phone's reported latitude and longitude (\texttt{latDeg\_phone}, \texttt{lngDeg\_phone}) for each epoch. From \texttt{ground\_truth.csv}, extract the NovAtel reference latitude and longitude (\texttt{latDeg\_gt}, \texttt{lngDeg\_gt}). Both are joined on the epoch identifier columns (\texttt{collectionName}, \texttt{phoneName}, \texttt{millisSinceGpsEpoch}).

\item \textbf{Extract error labels.} For each matched epoch from Step 1, convert both latitude/longitude pairs to ECEF coordinates, compute the difference vector, and rotate it into the local North-East frame at the reference location to obtain the error along the north direction (\texttt{error\_N\_m}), the error along the east direction (\texttt{error\_E\_m}), and the radial error (\texttt{radial\_error\_m}).

\item \textbf{Extract vehicle state and DOP.} From \texttt{ground\_truth.csv}, extract the speed of the vehicle (\texttt{speedMps}), the heading of the vehicle (\texttt{courseDegree}), the dilution of precision in horizontal and vertical dimensions (\texttt{hDop} and \texttt{vDop}), and the time elapsed since the first fix in seconds (\texttt{timeSinceFirstFixSeconds}).

\item \textbf{Aggregate signal records about satellite participation and geometry per epoch.} From \texttt{\{phone\_name\}\_GnssLog.txt}, group by the epoch identifier columns (\texttt{collectionName}, \texttt{phoneName}, \texttt{millisSinceGpsEpoch}) and compute the following per epoch:
\begin{itemize}
    \item \texttt{n\_sats\_used}: the total number of satellites participating in the fix, calculated by summing the \texttt{UsedInFix} flag.
    \item \texttt{avg\_cn0}: the average carrier-to-noise ratio in dB-Hz, measuring how strong each satellite signal is relative to background noise, from the \texttt{Cn0DbHz} column.
    \item \texttt{min\_cn0}: the minimum carrier-to-noise ratio across all tracked signals.
    \item \texttt{max\_cn0}: the maximum carrier-to-noise ratio across all tracked signals.
    \item \texttt{avg\_elevation}: the average elevation angle of the satellites in degrees, from the \texttt{ElevationDegrees} column.
    \item \texttt{min\_elevation}: the lowest elevation angle, corresponding to the satellite with the highest atmospheric delay risk.
    \item \texttt{max\_elevation}: the highest elevation angle, corresponding to the satellite with the clearest signal path.
    \item \texttt{avg\_azimuth}: the average direction from the receiver to each satellite in degrees, from the \texttt{AzimuthDegrees} column.
\end{itemize}

\item \textbf{Aggregate signal records about atmospheric delays and pseudorange uncertainty per epoch.} From \texttt{\{phone\_name\}\_derived.csv}, group by the epoch identifier columns (\texttt{collectionName}, \texttt{phoneName}, \texttt{millisSinceGpsEpoch}) and compute the following per epoch:
\begin{itemize}
    \item \texttt{avg\_iono}: the average ionospheric delay in meters, from the \texttt{ionoDelayM} column.
    \item \texttt{avg\_tropo}: the average tropospheric delay in meters, from the \texttt{tropoDelayM} column.
    \item \texttt{avg\_rawPrUnc}: the average pseudorange uncertainty in meters, computed from \texttt{rawPrUncM}. The pseudorange uncertainty is the receiver's self-reported confidence in its distance measurement to each satellite; higher values indicate noisier ranging conditions.
    \item \texttt{n\_signals}: the count of satellite signals for the epoch after quality filtering.
\end{itemize}

\item \textbf{Calculate the average satellite geometry cost per epoch according to \cite{pdf}.} Using \texttt{\{phone\_name\}\_derived.csv} and \texttt{baseline\_locations\_train.csv}, calculate \texttt{j\_avg}, defined as follows.

\smallskip
\noindent\textit{Let $\theta_{j,i}$ represent the angle between the line-of-sight vector 
from the receiver to satellite $i$ and the line-of-sight vector from the receiver to 
satellite $j$, where $i \neq j$. Following \cite{pdf} and \cite{park2001}, the four 
satellites forming the optimal geometric constellation are first selected by iteratively 
eliminating the most geometrically redundant satellite, the one with the highest 
individual geometry cost $J_j$. The selection of four satellites follows the quasi-optimal satellite selection algorithm of \cite{park2001}, which specifies four as the constellation size for optimal position fixation geometry:}
\begin{equation}
J_{j} = \sum_{\substack{i=1 \\ i \neq j}}^{4} \cos(2\theta_{j,i})
\label{eqn:onesatellitecost}
\end{equation}
\noindent\textit{The average satellite geometry cost over the four selected satellites is then:}
\begin{equation}
J_{\text{avg}} = \frac{1}{4} \sum_{j=1}^{4} J_j
\label{eqn:avgsatellitecost}
\end{equation}

\noindent The line-of-sight vectors are computed from the satellite ECEF positions in \texttt{\{phone\_name\}\_derived.csv} and the phone's reported receiver position from \texttt{baseline\_locations\_train.csv}, converted to ECEF coordinates.

\item \textbf{Join all feature tables.} Join all tables produced in Steps 1 to 6 on the epoch identifier columns (\texttt{collectionName}, \texttt{phoneName}, \texttt{millisSinceGpsEpoch}) to produce the final dataset, with one row per epoch containing all features and labels.

\end{enumerate}

\subsection{Cleaning}
The eight features derived from GnssLog records in Step 4, namely satellite participation, elevation, azimuth, and carrier-to-noise (CN0) metrics, were excluded due to inconsistent logging across devices and sessions. The analysis was then restricted to complete cases across \texttt{n\_signals}, \texttt{avg\_iono}, \texttt{avg\_rawPrUnc}, \texttt{avg\_tropo}, and \texttt{j\_avg}. Positioning error outliers, defined as observations falling beyond five interquartile ranges from the median in either the north or east error dimension, were then removed; this threshold was chosen to exclude only gross multipath failures and GNSS outages while retaining legitimate heavy-tailed observations. The final clean dataset comprised 73,537 observations across 20 columns: the three epoch identifiers, the reported and reference coordinates, the three error labels, and the ten features from Steps 3, 5, and 6.

The code and the resulting dataset are publicly available at: \url{https://www.kaggle.com/code/basantmounir/geofusion-dataset}.
\section{Implementation}
\label{sec:geo}

\begin{figure*}[!t]
\centering
\includegraphics[width=16.8cm]{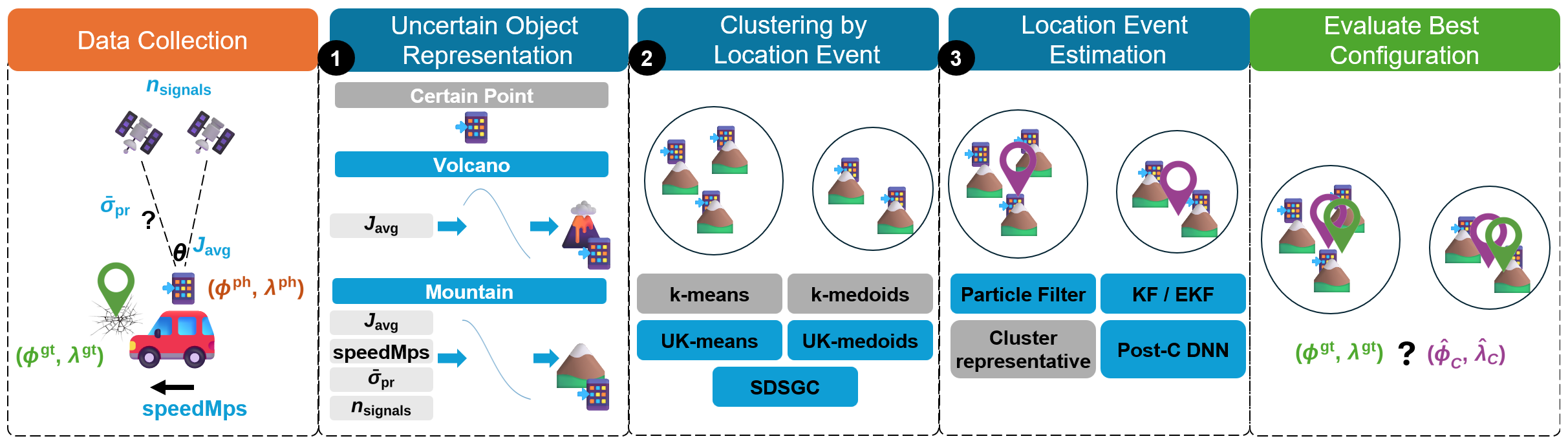}
\caption{The GeoFusion pipeline. After data collection, Stage~1 represents each phone-reported location (orange) as a certain point, a volcano function parameterized by $J_{\mathrm{avg}}$, or a mountain function parameterized by $J_{\mathrm{avg}}$, $\mathrm{speedMps}$, $n_{\mathrm{signals}}$, and $\bar{\sigma}_{\text{pr}}$ (blue); Stage~2 clusters these objects by location event (mountain case shown); and Stage~3 estimates each cluster's position using the cluster representative, KF, EKF, PF, or Post-C DNN, with certain baselines shown in gray. Each configuration's estimated location (purple) is evaluated against the ground truth (green), and the recommended configuration, UK-medoids with the mountain function followed by the Post-C DNN, is the one that embeds uncertainty at every stage and achieves the lowest error.}
\label{img:geo-framework}
\end{figure*}

\noindent
To achieve higher location event accuracy, the GeoFusion framework operates in three stages, illustrated in Fig.~\ref{img:geo-framework}.

\begin{enumerate}
    \item \textbf{Represent reported locations as uncertain objects.} GPS/GNSS positions reported by smartphones carry significant errors, typically ranging from 1 to 20 meters, stemming from atmospheric conditions, satellite geometry, multipath effects, and receiver clock drift. Rather than treating each reported position as a precise point, we model it as an \textit{uncertain object}, a probability distribution over all plausible true locations, weighted by the factors that contribute most to positioning error, which are selected empirically from the dataset in Section~\ref{sec:uncertain_objects}. Two uncertain object models are developed: the \textit{volcano function}, grounded in the satellite-geometry cost function of \cite{pdf}, and the \textit{mountain function}, derived directly from the statistical structure of the dataset.

    \item \textbf{Cluster uncertain objects by location event.} The aim is to group reported positions by the physical location event they correspond to, so that each cluster can later be fused into a single estimate. Because the input objects carry positional uncertainty, in addition to standard clustering algorithms, we consider uncertain clustering algorithms. Five algorithms are therefore evaluated: k-means~\cite{kmeans} and k-medoids~\cite{kmedoids} as point-based baselines, UK-means~\cite{ukmeans} and UK-medoids~\cite{ukmedoids} as their uncertainty-aware counterparts, and SDSGC~\cite{SDSGC} as a graph-based alternative.

    \item \textbf{Estimate the location event for each cluster.} Within each cluster, the reported positions are fused to produce a single location event estimate. The cluster representative (centroid or medoid, depending on the algorithm) serves as the primary GeoFusion estimate. For comparison, four additional estimators are evaluated against the same clusters: the Kalman filter~\cite{kalman1960new}, extended Kalman filter~\cite{kalman1961new}, particle filter~\cite{pf}, and a post-clustering deep neural network inspired by~\cite{siemuri2021}.
\end{enumerate}

Each stage is developed in detail in the remainder of this section.

\subsection{Locations as Uncertain Objects}
\label{sec:uncertain_objects}
\subsubsection{The Notion of Uncertain Objects}
In most data applications, data points are treated as exact, with no uncertainty in their values. However, to completely describe data points in contexts where uncertainty is a critical challenge, uncertainty must be part of the representation. A data point representation that specifies the multiple possible realizations along with their probabilities is termed an \textit{uncertain object} \cite{ukmedoids}.
\\
\\
\textit{\textbf{Uncertain Object Representation.}
Let \(O \subseteq \mathbb{R}^k \) be the domain that defines the possible realizations of the uncertain object. A probability density function is defined as \(f: O \rightarrow [0,1]\) where \(f\) is a transformation that maps each possible realization \(\mathbf{z} \in O\) to a probability \(f(\mathbf{z})\). An uncertain object is defined by the pair \(U = (O,f)\) fulfilling the following condition:}
\begin{equation}
\int_{\mathbf{z} \in O} f(\mathbf{z})\, \mathrm{d}\mathbf{z} = 1
\label{eqn:pdfcriteria}
\end{equation}
\textit{In this fashion, a certain object is defined as having a single possible realization with a probability of 1.}
\\

In this study, we consider two models of uncertain objects for the reported location: the isotropic \textit{volcano function} based on the error function in \cite{pdf}, and the direction-aware \textit{mountain function} implemented by statistical analysis of the dataset.

\subsubsection{The Volcano Function}
\noindent The \textit{volcano function} models a reported GPS location as an uncertain
object by associating it with a two-dimensional isotropic probability distribution over
the ground plane, parameterized by the geometry of the satellites that participated in
the position fix. Its construction follows \cite{pdf}: the satellite geometry is first
summarized by a scalar cost $J_{\mathrm{avg}}$, which is used to parameterize a
one-dimensional Gaussian over the radial distance error, which is then revolved about
the reported location to produce a radially symmetric two-dimensional distribution.

\textbf{Modeling the Distance Error Distribution.}
The state-of-the-art probability distribution function of the distance error used in
this work was proposed in \cite{pdf}. A normal distribution, whose mean and standard
deviation are functions of the average cost $J_{\mathrm{avg}}$ in
(\ref{eqn:avgsatellitecost}), is used to model the distance error.

\noindent
{\bf Distance Error Parameters.} \textit{Let $J_{\mathrm{avg}}$ be the average cost
associated with the fixation of a position. According to \cite{pdf}, the mean
$\mu(J_{\mathrm{avg}})$ and the standard deviation $\sigma(J_{\mathrm{avg}})$ of the
associated distance error distribution are defined as:}

\begin{align}
\mu(J_{\mathrm{avg}}) &= 0.605\, J_{\mathrm{avg}} + 1.945
\label{eqn:costmean}
\end{align}

\begin{align}
\sigma(J_{\mathrm{avg}}) &= 0.787\, J_{\mathrm{avg}} + 2.192
\label{eqn:coststd}
\end{align}

\noindent
{\bf Distance Error Distribution.} \textit{Given the average cost associated with the
fixation of the reported location, \cite{pdf} defines the distance error distribution
as:}

\begin{equation}
p(X \mid J_{\mathrm{avg}}) =
\frac{1}{\sigma\sqrt{2\pi}}
\exp\!\left(-\frac{(X - \mu)^2}{2\sigma^2}\right)
\label{eqn:gaussiandist}
\end{equation}

\noindent where $\mu = \mu(J_{\mathrm{avg}})$ and $\sigma = \sigma(J_{\mathrm{avg}})$
are given by (\ref{eqn:costmean}) and~(\ref{eqn:coststd}).

\textbf{From Distance Error Distribution to Volcano Distribution.}
The reported location is a two-dimensional vector defined by its latitude and longitude.
The probability distribution of the uncertain location is therefore expected to be
two-dimensional, with the output representing the probability density at each possible
location pair. Equation~(\ref{eqn:gaussiandist}), however, is one-dimensional: it models
the probability of the \textit{magnitude} of the positioning error, with no information
about direction.

Because the distance error distribution is isotropic, all points on the circumference
of radius $r$ from the reported location are equally probable. The two-dimensional
distribution is therefore obtained by revolving the one-dimensional distribution about
the reported location. Let $r \geq 0$ denote the radial distance and $\theta \in
[0, 2\pi)$ the angle. The unnormalized two-dimensional density is:

\begin{equation}
\tilde{g}(r, \theta) =
\frac{1}{\sigma\sqrt{2\pi}}
\exp\!\left(-\frac{(r - \mu)^2}{2\sigma^2}\right),
\quad r \geq 0
\label{eqn:unnorm}
\end{equation}

\noindent To satisfy (\ref{eqn:pdfcriteria}), $\tilde{g}$ is divided by its
volume $V$, the integral of the unnormalized density with area element
$r\,\mathrm{d}r\,\mathrm{d}\theta$:

\begin{equation}
V = \int_{0}^{2\pi}\!\int_{0}^{\infty}
\frac{1}{\sigma\sqrt{2\pi}}
\exp\!\left(-\frac{(r-\mu)^2}{2\sigma^2}\right)
r\;\mathrm{d}r\;\mathrm{d}\theta
\label{eqn:volumeintegral}
\end{equation}

\noindent Separating the angular integral yields a factor of $2\pi$. Letting
$\alpha = \mu/\sigma$, the radial integral evaluates in closed form as the first
moment of a truncated Gaussian:

\begin{equation}
V = 2\pi\bigl[\mu\,\Phi(\alpha) + \sigma\,\phi(\alpha)\bigr]
\label{eqn:Vclosed}
\end{equation}

\noindent where $\phi(\cdot)$ and $\Phi(\cdot)$ are the standard normal PDF and CDF
respectively, and $\alpha = \mu(J_{\mathrm{avg}})\,/\,\sigma(J_{\mathrm{avg}})$. Since
$\mu(J_{\mathrm{avg}}) > 0$ for all $J_{\mathrm{avg}} \in [-1.7,\, 0.8]$,
$V > 0$ is guaranteed.

{\bf Volcano Function.} \textit{Let $J_{\mathrm{avg}}$ be the average cost associated
with the geometry of the satellites during the fixation of the reported location.
Assuming the reported location is at the origin, the normalized volcano distribution
in polar coordinates is:}

\begin{equation}
g(r,\, J_{\mathrm{avg}}) =
\frac{1}{V\,\sigma\sqrt{2\pi}}
\exp\!\left(-\frac{(r - \mu)^2}{2\sigma^2}\right),
\quad r \geq 0
\label{eqn:volcanofunction}
\end{equation}

\noindent \textit{where $r$ is the radial distance from the reported location,
$\mu = \mu(J_{\mathrm{avg}})$, $\sigma = \sigma(J_{\mathrm{avg}})$, and $V$ are given
by (\ref{eqn:costmean}), (\ref{eqn:coststd}), and~(\ref{eqn:Vclosed})
respectively. The volcano distribution for any reported location is obtained by
shifting the origin to the reported location coordinates.}

\begin{figure}[!t]
\centering
\subfloat[Error Distribution]{%
  \includegraphics[width=3.7cm]{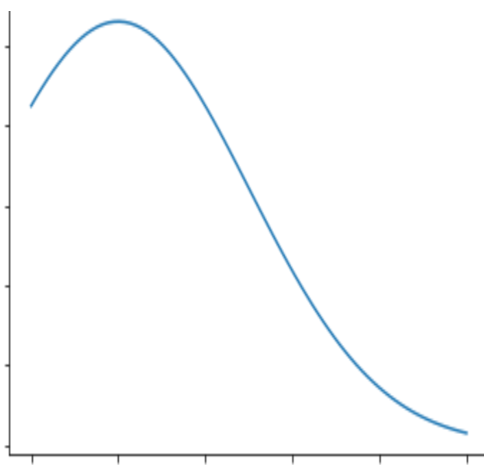}%
  \label{fig:errormagnormal}%
}\qquad
\subfloat[Volcano Function]{%
  \includegraphics[width=3.7cm]{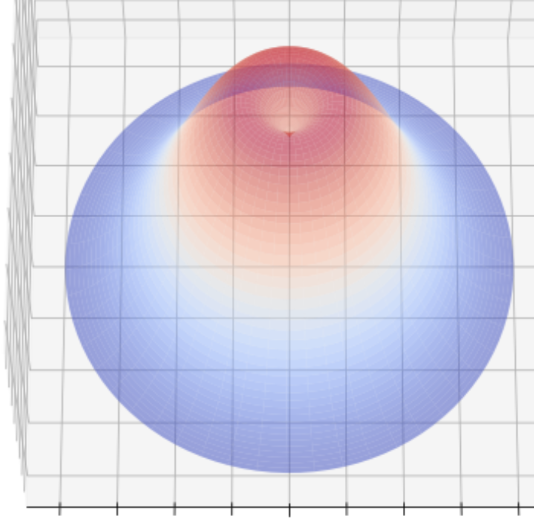}%
  \label{fig:volcanofunctiongraph}%
}
\caption{In \protect\subref{fig:errormagnormal}, the distance error distribution for a
particular reported location is modeled by (\ref{eqn:gaussiandist}). The
portion of the distribution over the domain $[0, +\infty)$ is revolved about the
vertical axis to produce the two-dimensional volcano function shown in
\protect\subref{fig:volcanofunctiongraph}, normalized by $V$ in
(\ref{eqn:Vclosed}).}
\label{fig:volcanographs}
\end{figure}

\subsubsection{The Mountain Function}
The mountain function is an empirically derived uncertain object representation that
models the probability distribution of the true location given a reported GPS fix.
Unlike the volcano function, which adopts the Gaussian error model of \cite{pdf}
parameterized by satellite geometry alone, the mountain function is built directly from
the statistical structure of the dataset and is fit separately in the north and east
directions. Empirical analysis reveals that GPS positioning errors along both the north
and east directions follow heavy-tailed Student-$t$ distributions. The spread of each
marginal is modeled as a function of four receiver observables, satellite geometry
cost, vehicle speed, number of visible signals, and pseudorange uncertainty
($\overline{\sigma}_{\mathrm{pr}}$, corresponding to \texttt{avg\_rawPrUnc} in the
dataset), via per-drive multivariate regression. Since the mountain function
represents the distribution of the true location rather than the error, the fitted
error distributions are reflected across the origin. The result is a fully empirical,
per-observation uncertain object that captures the asymmetric, heavy-tailed nature of
GPS noise while remaining computable in real time from standard receiver outputs.

\begin{figure}[!t]
\centering
\includegraphics[width=8.4cm]{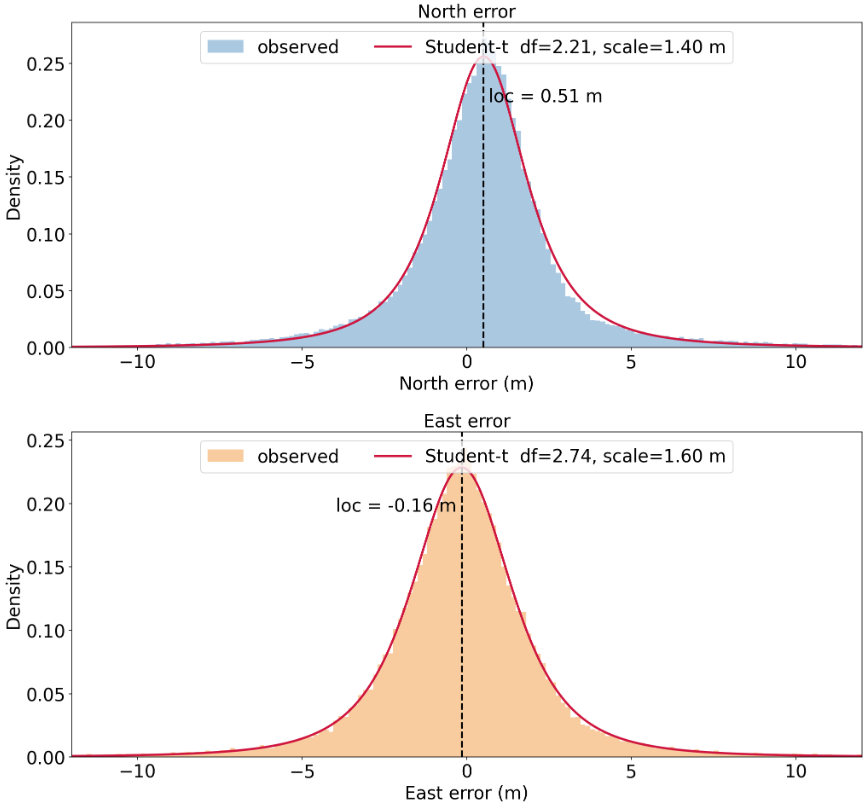}
\caption{The Student-$t$ distributions of errors along the north and east directions.}
\label{fig:error-dist-mountain}
\end{figure}

\textbf{Fitting the Functional Form.} To identify the functional form of the uncertain
object, the distribution of the error is studied empirically. Along the north
direction, the error distribution is bell-shaped and heavy-tailed, which mirrors a
Student-$t$ distribution with 2.209 degrees of freedom. For the same reasons, the east
direction can also be modeled as a Student-$t$ distribution, with 2.743 degrees of
freedom. Both distributions are shown in Fig.~\ref{fig:error-dist-mountain}.

\textbf{Per-point Parameters of the Mountain Function.} The next step is to define
the location and scale of the distributions according to the error-contributing factors
of each reported location. Reported locations with larger errors should have uncertain
objects with larger scale. To identify factors that significantly contribute to the
error, the correlation of each feature is measured with the error along the east and
north directions as well as the overall magnitude of the error, which we refer to as
radial error.

\begin{figure}[!t]
\centering
\includegraphics[width=8.4cm]{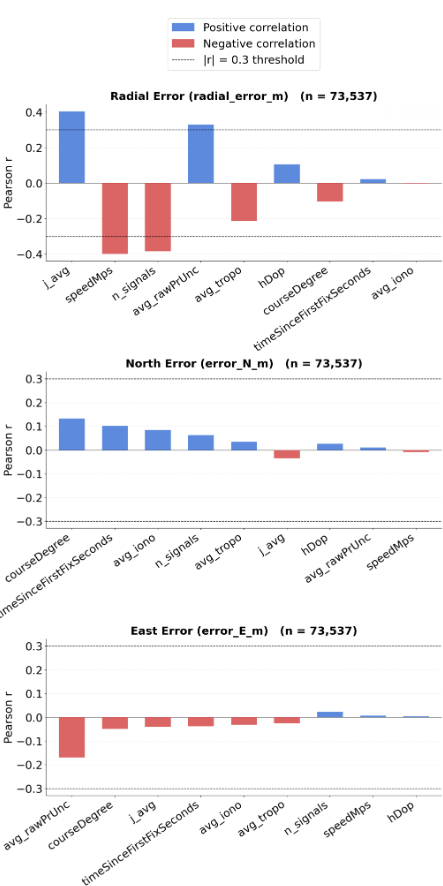}
\caption{The correlations of the features with the overall magnitude of error, the
radial error, as well as errors along the north and east components. Four features,
$J_{\mathrm{avg}}$, $\mathrm{speedMps}$, $n_{\mathrm{signals}}$, and
$\overline{\sigma}_{\mathrm{pr}}$, exceed the $|r| = 0.3$ threshold on the radial
error and are used to model the scale and location of the Student-$t$ uncertain
object.}
\label{fig:feature-importance}
\end{figure}

We set a minimum threshold of correlation of 0.3 to consider the features in the model
of the Student-$t$ distribution location and scale. The results are shown in
Fig.~\ref{fig:feature-importance}. The radial error shows promising results with the average
cost ($J_{\mathrm{avg}}$), speed ($\mathrm{speedMps}$), number of signals
($n_{\mathrm{signals}}$), and average pseudorange uncertainty
($\overline{\sigma}_{\mathrm{pr}}$) showing significant correlation. Along the
individual north and east components, no significant correlation was observed.
These four features are used as independent variables throughout the remainder of this
section.

With the important variables identified, we fit regression models of the scale and
location along both the north and east directions. Like \cite{pdf}, we first bin the
location events by drive, since each drive represents a distinct measurement
campaign. Next, for each bin, we calculate the means of the four selected features as
the independent variables.

To model the standard deviation, we calculate the error standard deviation in each bin
as the dependent variable and fit a multivariate linear regression model separately for
the north and east directions. Since the marginal distributions are Student-$t$ rather
than Gaussian, the empirical standard deviation $\hat{\sigma}$ is converted to the
Student-$t$ scale parameter $s$ via $s = \hat{\sigma}\sqrt{(\nu - 2)/\nu}$.
The resulting per-observation scale parameters, with $R^2$ values of $0.74$ and $0.75$
for the north and east directions respectively, are:

\begin{equation}
\begin{split}
s_N = \Big( 7.398 - 0.090\, J_{\mathrm{avg}} - 0.103\, \mathrm{speedMps} \\
      - 0.151\, n_{\mathrm{signals}} + 0.069\, \overline{\sigma}_{\mathrm{pr}} \Big)
      \sqrt{\frac{\nu_N - 2}{\nu_N}}
\end{split}
\label{eqn:scale_N}
\end{equation}

\begin{equation}
\begin{split}
s_E = \Big( 7.612 + 0.031\, J_{\mathrm{avg}} - 0.099\, \mathrm{speedMps} \\
      - 0.116\, n_{\mathrm{signals}} + 0.012\, \overline{\sigma}_{\mathrm{pr}} \Big)
      \sqrt{\frac{\nu_E - 2}{\nu_E}}
\end{split}
\label{eqn:scale_E}
\end{equation}

\noindent where $\nu_N = 2.209$, $\nu_E = 2.743$, $\mathrm{speedMps}$ is in m/s,
$\overline{\sigma}_{\mathrm{pr}}$ is in meters, and $J_{\mathrm{avg}}$ and
$n_{\mathrm{signals}}$ are dimensionless.

The same process is repeated for the location parameter, producing regression models
with $R^2 = 0.29$ for the east direction and $R^2 = 0.10$ for the north direction.
Because the $R^2$ values are low, fixed location parameters are used instead:
$\mu_N = 0.512$ m and $\mu_E = -0.156$ m, estimated directly from the Student-$t$
distributions fitted to the dataset.

\textbf{The Mountain Function.} The mountain function defines the probability
distribution of the true location of an uncertain object given a reported GPS fix.
Since the mountain function models the distribution of the true location relative to
the reported fix, rather than the error itself, the error distribution is reflected
across the origin. Concretely, a northward error bias of $\mu_N = 0.512$ m implies the
true location is expected $0.512$ m south of the reported fix, and a westward error
bias of $\mu_E = -0.156$ m implies the true location is expected $0.156$ m east of
the reported fix. The fixed parameters used in the mountain function therefore reflect
these signs: after reflection, the location parameters entering the mountain function
are $-\mu_N = -0.512$ m along the north direction and $-\mu_E = +0.156$ m along the
east direction, as summarized in Table~\ref{tab:mountain_params}.

The mountain function is constructed as the product of two independent univariate
Student-$t$ marginals, one along the north direction and one along the east direction,
justified by the near-zero correlation between north and east errors (Pearson
$r = -0.060$, Spearman $r = -0.033$). The joint mountain function is defined as:

\begin{equation}
g(x_N, x_E) = f_N(x_N;\, \nu_N,\, \mu_N,\, s_N)\, f_E(x_E;\, \nu_E,\, \mu_E,\, s_E)
\label{eqn:mountain}
\end{equation}

\noindent where $x_N$ and $x_E$ are the displacements in meters from the reported GPS
location along the north and east directions respectively, and each marginal density
takes the standard Student-$t$ form:

\begin{equation}
f(x;\, \nu,\, \mu,\, s) =
\frac{\Gamma\!\left(\frac{\nu+1}{2}\right)}
     {\Gamma\!\left(\frac{\nu}{2}\right) \sqrt{\nu\pi}\, s}
\left(1 + \frac{(x - \mu)^2}{\nu\, s^2}\right)^{-\frac{\nu+1}{2}}
\label{eqn:student_t_marginal}
\end{equation}

\noindent The fixed parameters of the two marginals are summarized in Table
\ref{tab:mountain_params}, and the per-observation scale parameters $s_N$ and $s_E$
are defined by (\ref{eqn:scale_N}) and~(\ref{eqn:scale_E}).

\begin{table}[!t]
\renewcommand{\arraystretch}{1.3}
\caption{Fixed parameters of the mountain function marginals}
\label{tab:mountain_params}
\centering
\begin{tabular}{lccc}
\hline
\textbf{Direction} & \textbf{df} ($\nu$) & \textbf{loc} ($\mu$, m) & \textbf{Scale} \\
\hline
North & 2.209 & $-0.512$ & Eq.~(\ref{eqn:scale_N}) \\
East  & 2.743 & $+0.156$ & Eq.~(\ref{eqn:scale_E}) \\
\hline
\end{tabular}
\end{table}

\noindent Since both $f_N$ and $f_E$ are proper probability densities integrating to
unity over $\mathbb{R}$, their product integrates to unity over $\mathbb{R}^2$,
satisfying the uncertain object normalization condition in
(\ref{eqn:pdfcriteria}) with no additional normalization constant required.

\subsection{Clustering Locations by Location Event}
\label{sec:clustering}

The second stage of the GeoFusion framework groups the uncertain objects produced in
Section~\ref{sec:uncertain_objects} into clusters by location event, so that the
observations from each cluster can subsequently be fused into a single position
estimate. Two algorithmic paradigms are considered for this task. The first is
partitional clustering, in which the data are directly divided into \(k\) groups by
optimizing an objective function over cluster assignments. Within this paradigm, four
algorithms are evaluated: k-means \cite{kmeans} and k-medoids \cite{kmedoids} as
point-based baselines, and UK-means \cite{ukmeans} and UK-medoids \cite{ukmedoids} as
their uncertainty-aware counterparts. The second
paradigm is graph-based clustering, in which a structured affinity graph is learned
over the data and cluster membership is extracted from its topology. SDSGC
\cite{SDSGC} is evaluated as the representative of this paradigm, selected from a
family of recent graph-based methods \cite{patcog2026, neucom130571, sigpro110144}.

\subsubsection{Partitional Clustering}
Typically, partitional clustering algorithms divide the data into \(k\) groups of
similar objects by optimizing an objective function over cluster assignments, with each
object belonging to exactly one cluster. Algorithms in this paradigm consist of two
components: a cluster representative \(\rho\) and a similarity metric \(\delta(\mathbf{x},\mathbf{y})\).
A cluster representative embodies a collective property of the group and could be an
estimate calculated from the group or a group member. To measure the degree of
belonging of an object to a certain group, a similarity metric is used. This metric
specifies the objective function to be minimized for optimal clustering.

Well-known partitional clustering algorithms, such as k-means and Partitioning Around
Medoids (PAM), have uncertain counterparts, namely UK-means \cite{ukmeans} and
UK-medoids \cite{ukmedoids}. Both algorithms use distance-based similarity metrics to
assign objects to clusters. Novel uncertain algorithms with no certain counterparts
have also been devised, most notably MMVar \cite{mmvar} and Uncertain Centroid-based
Partitional Clustering (UCPC) \cite{ucpc}. Unlike UK-means and UK-medoids, MMVar
clusters by variance, not proximity. Furthermore, the cluster representative of UCPC
is not analytically computable \cite{ucpc}, rendering the algorithm's output itself
intractable. Therefore, out of the four algorithms, UK-means and UK-medoids are the
candidates considered for the GeoFusion framework. Both algorithms share the same
structure as shown in Algorithm~\ref{algo:ukmeans}.

\begin{figure}[!t]
\removelatexerror
\begin{algorithm}[H]
\SetAlgoLined
{\bf Input:} A set of \(n\) uncertain objects \(S = \{ U_1, U_2, U_3,\ldots,U_n\}\) to be clustered and the number of clusters to be formed \(k\) \\
{\bf Output:} Clusters \(\omega\) and their respective representatives \(\eta\)

{\bf Initialize:} All cluster representatives \(\rho \in \eta\) using k-means\texttt{++} seeding \cite{kmeansplusplus} with \texttt{n\_init} independent restarts; the run with the lowest objective value is retained.

\While{clusters change}{
  \For{all \(U \in S\)}{
    assign each object \(U\) to the cluster \(C\) where \(\delta(U, \rho)\) is minimum \\
  }
  \For{all \(\rho \in \eta\)}{
    recompute the cluster representative \(\rho\) of cluster \(C\) using (\ref{eqn:centroidukmeans}) for UK-means or (\ref{eqn:centroidukmedoids}) for UK-medoids based on the new assignments from the previous loop
  }
}
\caption{Uncertain Proximity-based Partitional Algorithms}
\label{algo:ukmeans}
\end{algorithm}
\end{figure}

\noindent\textbf{Cluster Representatives.} The centroid, denoted \(\boldsymbol{\rho}_C\), is the
representative of a UK-means cluster. It is the average of the means of the uncertain
objects:
\begin{equation}
\boldsymbol{\rho}_C = \frac{1}{n_C} \sum_{U \in C} \boldsymbol{\mu}(O)
\label{eqn:centroidukmeans}
\end{equation}
where \(O\) is the random variable of the uncertain object \(U\) and \(n_C\) is the number of objects in cluster \(C\). 
Since $\boldsymbol{\rho}_C$ is defined solely in terms of the means $\boldsymbol{\mu}(O)$ of the 
cluster members in (\ref{eqn:centroidukmeans}), the per-object uncertainty spread 
$\mathrm{tr}(\Sigma_i)$ enters the assignment step only as an additive constant that is invariant to cluster membership. The UK-means assignment therefore reduces to
\begin{equation}
    \arg\min_{C} \|\boldsymbol{\mu}_i - \boldsymbol{\rho}_C\|^2,
    \label{eq:ukmeans_reduce}
\end{equation}
which is independent of the uncertainty model. Consequently, the 
\textit{volcano} and \textit{mountain} functions yield identical 
UK-means partitions for any given $k$, a result confirmed empirically in Section~\ref{sec:experiments}.

The representative
of a UK-medoids cluster, the medoid \(\rho_M\), is the cluster member with the minimum
total expected distance from all other members:
\begin{equation}
\rho_M(C) = \underset{U \in C}{\arg\min} \left( \sum_{U' \in C - \{U\}} \delta(U', U) \right)
\label{eqn:centroidukmedoids}
\end{equation}
where \(\delta(U',U)\) is the expected distance between two uncertain objects, defined
next.

\noindent\textbf{Similarity Metrics.} Both UK-means and UK-medoids require computing
distances involving uncertain objects. The \textit{expected distance}, devised in
\cite{ukmedoids}, is defined as follows. Let \(U_i = (O_i, f_i)\) and
\(U_j = (O_j, f_j)\) be uncertain objects with realization spaces
\(O_i \subseteq \mathbb{R}^k\) and \(O_j \subseteq \mathbb{R}^k\), and probability
distribution functions \(f_i: O_i \rightarrow [0,1]\) and
\(f_j: O_j \rightarrow [0,1]\), respectively. The expected distance
\(\delta(U_i, U_j)\) is:
\begin{equation}
\mathrm{E}[\mathrm{dist}(U_i,U_j)] = \int_{\mathbf{x} \in O_i} \int_{\mathbf{y} \in O_j} \mathrm{dist}(\mathbf{x},\mathbf{y})\, f_i(\mathbf{x})\, f_j(\mathbf{y})\, \mathrm{d}\mathbf{x}\, \mathrm{d}\mathbf{y}
\label{eqn:exp_distance}
\end{equation}
where \(\mathrm{dist}(\mathbf{x},\mathbf{y})\) is the distance function. The similarity metric
of UK-means is a special case adapted for a certain centroid \(\boldsymbol{\rho}_C\):
\begin{equation}
\mathrm{E}[\mathrm{dist}(U,\boldsymbol{\rho}_C)] = \int_{\mathbf{x} \in O} \mathrm{dist}(\mathbf{x},\boldsymbol{\rho}_C)\, f(\mathbf{x})\, \mathrm{d}\mathbf{x}
\label{eqn:exp_distance_c_u}
\end{equation}

\noindent\textbf{Expected Distance Simplification.} Instead of directly computing the
integrals in (\ref{eqn:exp_distance}) and (\ref{eqn:exp_distance_c_u}), the expected
distances can be reformulated for efficient implementation. For UK-medoids, there is
no closed form for the expected Euclidean distance between two independent Gaussians
with direction-specific scales \(\sigma_N\) and \(\sigma_E\). The expected distance
matrix is therefore estimated via Monte Carlo sampling with \texttt{n\_samples} paired
draws per object pair, matching the methodology of \cite{ukmedoids}.

\subsubsection{Graph-Based Clustering}

Graph-based clustering is a distinct paradigm in which a similarity structure over the
data is first learned as an affinity graph, and cluster membership is then extracted
from the topology of that graph rather than by directly optimizing a partition. Unlike
in partitional algorithms, the number of clusters is not enforced during optimization but
instead emerges from the graph structure, typically as connected components or via
spectral decomposition of the graph Laplacian.

\noindent\textbf{Structured Doubly Stochastic Graph-Based Clustering.} Recent
graph-based clustering methods have focused on learning a doubly stochastic affinity
matrix \(W\), which is non-negative, with rows and columns each summing to one, to
simultaneously suppress inter-cluster connections and enhance intra-cluster connections
\cite{SDSGC, patcog2026,  sigpro110144, neucom130571}. Among these, SDSGC
\cite{SDSGC} addresses three limitations of prior work: dependence on a predefined
graph, the mismatched multi-stage pipeline of graph learning followed by separate
spectral embedding and k-means assignment, and the use of the von Neumann successive
projection (VNSP) lemma, which converges only to a feasible rather than optimal
solution. SDSGC jointly learns a structured doubly stochastic graph directly from the
data, providing cluster indicators without a separate assignment step. Given a data
matrix \(X \in \mathbb{R}^{d \times n}\) and a predefined affinity matrix \(A\),
SDSGC solves:
\begin{equation}
\begin{split}
\min_{\substack{W \geq 0,\; W\mathbf{1}=\mathbf{1},\; W=W^\top}}
& \sum_{i,j} \|\mathbf{x}_i - \mathbf{x}_j\|_2^2\, w_{ij} + \gamma \|W\|_F^2 \\
& + 2\eta\, \mathrm{tr}(F^\top L_W F)
\end{split}
\label{eqn:sdsgc}
\end{equation}
subject to \(\mathrm{rank}(L_W) = n - k\) and \(\mathrm{tr}(W) = k\), where
\(L_W = D_W - W\) is the graph Laplacian of \(W\), \(F \in \mathbb{R}^{n \times k}\)
is the spectral embedding matrix whose columns are the \(k\) smallest eigenvectors of
\(L_W\), \(\gamma\) controls the sparsity of \(W\), and \(\eta\) balances the spectral
embedding term. The rank constraint enforces exactly \(k\) connected components in
\(W\), from which cluster labels are read directly. The optimization is solved via an
augmented Lagrangian multiplier (ALM) scheme that enforces all doubly stochastic
conditions simultaneously, as summarized in Algorithm~\ref{algo:sdsgc}.

\begin{figure}[!t]
\removelatexerror
\begin{algorithm}[H]
\SetAlgoLined
{\bf Input:} Data matrix \(X \in \mathbb{R}^{d \times n}\), number of clusters \(k\), regularization \(\gamma\), embedding weight \(\eta\), number of neighbors \(n_{\mathrm{nbr}}\) \\
{\bf Output:} Cluster labels \(\mathbf{y} \in \mathbb{Z}^n\), learned graph \(W \in \mathbb{R}^{n \times n}\)

{\bf Initialize:} Construct initial affinity \(W\) via a self-tuned Gaussian kernel over the \(n_{\mathrm{nbr}}\)-nearest-neighbor graph; normalize to doubly stochastic via Sinkhorn-Knopp. Set Lagrange multiplier \(\lambda = 0\), penalty \(\mu = 1\). Compute \(F\) as the \(k\) smallest eigenvectors of \(L_W\).

\While{not converged}{
  Compute \(Z = \mu W^\top - \tfrac{1}{2}\lambda + \tfrac{1}{2}\lambda^\top - \tfrac{1}{2}D_X - \tfrac{1}{2}\eta\, D_F\) where \(D_X\), \(D_F\) are pairwise squared distance matrices of \(X\) and \(F\) \\
  Update each row \(i\) of \(W\) by projecting \(\frac{Z_{i,\mathcal{N}(i)}}{\gamma + \mu}\) onto the probability simplex \\
  Update Lagrange multiplier: \(\lambda \leftarrow \lambda + \mu(W - W^\top)\) \\
  Update penalty: \(\mu \leftarrow \rho\,\mu\) \\
  Symmetrize: \(W \leftarrow \tfrac{1}{2}(W + W^\top)\) \\
  Update \(F\) as the \(k\) smallest eigenvectors of \(L_W\) \\
  \lIf{\(\mathrm{rank}(L_W) = n - k\) and \(W\) is doubly stochastic}{\textbf{break}}
  Adapt \(\eta\) based on spectral gap around the \(k\)-th eigenvalue of \(L_W\)
}
Extract cluster labels \(\mathbf{y}\) from the connected components of \(W\)
\caption{Structured Doubly Stochastic Graph-Based Clustering (SDSGC) \cite{SDSGC}}
\label{algo:sdsgc}
\end{algorithm}
\end{figure}

\noindent\textbf{Uncertainty in Graph-Based Clustering.} The graph-based clustering
works cited above \cite{SDSGC, patcog2026, neucom130571, sigpro110144}
express uncertainty through the soft affinity weights of the doubly stochastic matrix
\(W\), reflecting the degree to which two points are likely to belong to the same
cluster. This is distinct from the uncertain object formalism of \cite{ukmeans,
ukmedoids} adopted in the partitional paradigm, where uncertainty is embedded at the
level of the data point itself, modeling each observation as a probability distribution
over possible true locations.

\subsection{Location Event Estimation}

The third stage of the GeoFusion framework fuses the observations within each cluster
into a single position estimate for the corresponding location event. The cluster
representative produced by the preceding clustering step, either the centroid or medoid
of the reported phone positions depending on the algorithm, serves as the primary
GeoFusion estimate and as the initialization point for all downstream estimators.

Four estimators are evaluated against the same clusters: the Kalman filter, the
extended Kalman filter, the particle filter, and a post-clustering deep neural network.
All four share the same input structure: observations are grouped by cluster label,
sorted in temporal order within each cluster, and processed independently per cluster.
The output of each estimator is a single refined position estimate
$(\hat{\phi}_C, \hat{\lambda}_C)$ per cluster $C$, where $\hat{\phi}$ and
$\hat{\lambda}$ denote estimated latitude and longitude in degrees.

\subsubsection{Kalman Filter}
\label{sec:kf}

The Kalman filter~\cite{kalman1960new} is a recursive Bayesian estimator that maintains
a Gaussian belief over the state and updates it optimally under linear-Gaussian
assumptions. Within a cluster, all observations correspond to the same fixed location
event observed repeatedly under varying signal conditions. The motion model is therefore
a \textit{static target}: the true position does not change between observations, and
any apparent variation is attributed to measurement noise.

\textbf{State and Motion Model.}
The state vector is the two-dimensional position in degrees,
$\mathbf{x} = [\phi,\, \lambda]^\top$. The transition model is:
\begin{equation}
\mathbf{x}_{t+1} = F\,\mathbf{x}_t + \mathbf{w}_t, \quad \mathbf{w}_t \sim \mathcal{N}(\mathbf{0}, Q)
\label{eqn:kf_transition}
\end{equation}
where $F = I_2$ (identity, static target) and $Q = q\,I_2$, with scalar $q$
controlling the assumed residual positional drift between observations. The measurement
model is:
\begin{equation}
\mathbf{z}_t = H\,\mathbf{x}_t + \mathbf{v}_t, \quad \mathbf{v}_t \sim \mathcal{N}(\mathbf{0}, R_t)
\label{eqn:kf_measurement}
\end{equation}
where $H = I_2$ and $\mathbf{z}_t = [\phi_t^{\mathrm{ph}},\, \lambda_t^{\mathrm{ph}}]^\top$
is the reported phone position at epoch $t$. The measurement noise covariance is
$R_t = \overline{\sigma}_{\mathrm{pr},t}^2\, I_2$, where the per-epoch pseudorange
uncertainty $\overline{\sigma}_{\mathrm{pr},t}$ (\texttt{avg\_rawPrUnc}) is used
directly as the measurement noise standard deviation, assigning less weight to noisier
epochs with poorer signal quality.

\textbf{Initialization.}
The filter is initialized at the cluster representative $\boldsymbol{\rho}_C = (\phi_C, \lambda_C)$,
the centroid or medoid of the reported phone positions within the cluster, depending on the clustering algorithm. The initial state
covariance is $P_0 = h_{\mathrm{DOP},0}^2\, I_2$, where $h_{\mathrm{DOP},0}$ is the
horizontal dilution of precision (\texttt{hDop}) of the first cluster member in
temporal order.

\textbf{Filter Equations.}
At each epoch $t$, the predict step propagates the state and covariance forward:
\begin{equation}
\mathbf{x}_{t|t-1} = F\,\mathbf{x}_{t-1|t-1}, \quad
P_{t|t-1} = F\,P_{t-1|t-1}\,F^\top + Q
\label{eqn:kf_predict}
\end{equation}
The update step incorporates the new measurement:
\begin{equation}
S_t = H\,P_{t|t-1}\,H^\top + R_t, \quad
K_t = P_{t|t-1}\,H^\top S_t^{-1}
\label{eqn:kf_gain}
\end{equation}
\begin{equation}
\mathbf{x}_{t|t} = \mathbf{x}_{t|t-1} + K_t\,(\mathbf{z}_t - H\,\mathbf{x}_{t|t-1})
\label{eqn:kf_update_x}
\end{equation}
\begin{equation}
P_{t|t} = (I_2 - K_t\,H)\,P_{t|t-1}
\label{eqn:kf_update_P}
\end{equation}
where $S_t$ is the innovation covariance and $K_t$ is the Kalman gain. After
processing all $n_C$ members of cluster $C$ in temporal order, the final state
$\mathbf{x}_{n_C|n_C}$ is taken as the refined position estimate
$(\hat{\phi}_C^{\mathrm{KF}},\, \hat{\lambda}_C^{\mathrm{KF}})$.

\subsubsection{Extended Kalman Filter}
\label{sec:ekf}

The linear Kalman filter operates in degree space, where the measurement noise
covariance $R_t$ is expressed in degree$^2$ units. However, $\overline{\sigma}_{\mathrm{pr},t}$
and $h_{\mathrm{DOP}}$ are both given in meters: one degree of latitude is
approximately $111{,}133$ m, while one degree of longitude depends on latitude.
Applying meter-scale noise values directly in degree space introduces a distortion that
makes the covariance matrix physically uninterpretable.

The \textit{extended} Kalman filter~\cite{kalman1961new} used here resolves this by
converting all positions to a local North-East metric frame before filtering, so that
$R_t$ and $P_0$ are expressed in meters$^2$ and directly commensurate with
$\overline{\sigma}_{\mathrm{pr},t}$ and $h_{\mathrm{DOP}}$. The state vector,
transition model, and measurement model are structurally identical to those in
Section~\ref{sec:kf}, but redefined in meters.

\textbf{Coordinate Transformation.}
Each cluster defines a local Cartesian frame centered at $(\phi_C, \lambda_C)$. The
phone position at epoch $t$ is converted to local (North, East) displacements in
meters:
\begin{equation}
N_t = (\phi_t^{\mathrm{ph}} - \phi_C)\, m_{\mathrm{lat}}(\phi_C)
\label{eqn:ekf_N}
\end{equation}
\begin{equation}
E_t = (\lambda_t^{\mathrm{ph}} - \lambda_C)\, m_{\mathrm{lon}}(\phi_C)
\label{eqn:ekf_E}
\end{equation}
where $m_{\mathrm{lat}}$ and $m_{\mathrm{lon}}$ are the WGS84 meters-per-degree
conversion factors at latitude $\phi_C$, defined in (\ref{eqn:degree_lat})
and~(\ref{eqn:degree_lon}). The state is initialized at $\mathbf{x}_0 = \mathbf{0}$
with $P_0 = h_{\mathrm{DOP},0}^2\,I_2$ meters$^2$, and the measurement noise is
$R_t = \overline{\sigma}_{\mathrm{pr},t}^2\,I_2$ meters$^2$, applied directly without
unit conversion.

\noindent
{\bf Converting Latitude and Longitude to Meters.} \textit{For a given location, let \(\phi\) be its latitude in degrees. The meters per degree of latitude \(m_{\mathrm{lat}}(\phi)\) and of longitude \(m_{\mathrm{lon}}(\phi)\) at that location are respectively defined as:}

\begin{align}
\begin{split}
m_{\mathrm{lat}}(\phi) &= 111{,}132.92 - 559.82\cos(2\phi) \\
&\quad + 1.175\cos(4\phi) - 0.0023\cos(6\phi)
\end{split}
\label{eqn:degree_lat}
\end{align}

\begin{align}
\begin{split}
m_{\mathrm{lon}}(\phi) &= 111{,}412.84\cos\phi - 93.5\cos(3\phi) \\
&\quad + 0.118\cos(5\phi)
\end{split}
\label{eqn:degree_lon}
\end{align}

Since both the transition and measurement functions are linear in this local frame,
the EKF equations are identical to (\ref{eqn:kf_predict})
through~(\ref{eqn:kf_update_P}). After processing all cluster members, the final
local-frame estimate $(\hat{N}_C, \hat{E}_C)$ is converted back to degrees:
\begin{equation}
\hat{\phi}_C^{\mathrm{EKF}} = \phi_C + \frac{\hat{N}_C}{m_{\mathrm{lat}}(\phi_C)}
\label{eqn:ekf_back_lat}
\end{equation}
\begin{equation}
\hat{\lambda}_C^{\mathrm{EKF}} = \lambda_C + \frac{\hat{E}_C}{m_{\mathrm{lon}}(\phi_C)}
\label{eqn:ekf_back_lon}
\end{equation}

The distinction between the KF and EKF estimators therefore lies not in the filter
equations but in the coordinate frame: the KF filters in degree space with meter-scale
noise values applied as-is, while the EKF first maps observations to a metric frame
where noise values are physically meaningful, filters there, and maps the result back.

\subsubsection{Particle Filter}
\label{sec:pf}

The particle filter~\cite{pf} is a Sequential Monte Carlo estimator that approximates
the posterior distribution over the state as a weighted set of samples, or particles,
without assuming Gaussianity. This makes it applicable in principle to any likelihood
function and any noise distribution.

The particle filter operates in the same local North-East metric frame defined in
Section~\ref{sec:ekf}, with the same static target motion model. At each epoch, the
filter maintains a set of $P$ particles $\{\mathbf{x}_t^{(p)}\}_{p=1}^P$, each
representing a candidate position in meters, together with associated normalized
weights $\{w_t^{(p)}\}_{p=1}^P$.

\textbf{Initialization.}
Particles are drawn from the Gaussian prior centered at the cluster representative:
\begin{equation}
\mathbf{x}_0^{(p)} \sim \mathcal{N}\!\left(\mathbf{0},\; h_{\mathrm{DOP},0}^2\,I_2\right), \quad p = 1,\ldots,P
\label{eqn:pf_init}
\end{equation}
with uniform initial weights $w_0^{(p)} = 1/P$.

\textbf{Propagation.}
At each epoch $t$, particles are propagated according to the static target motion
model:
\begin{equation}
\mathbf{x}_t^{(p)} = \mathbf{x}_{t-1}^{(p)} + \boldsymbol{\epsilon}_t^{(p)}, \quad \boldsymbol{\epsilon}_t^{(p)} \sim \mathcal{N}(\mathbf{0},\, q\,I_2)
\label{eqn:pf_propagate}
\end{equation}

\textbf{Weighting.}
Each particle is weighted by the Gaussian measurement likelihood at epoch $t$:
\begin{equation}
\tilde{w}_t^{(p)} \propto \exp\!\left(\frac{-\|\mathbf{x}_t^{(p)} - \mathbf{z}_t\|^2}{2\,\overline{\sigma}_{\mathrm{pr},t}^2}\right)
\label{eqn:pf_weight}
\end{equation}
where $\mathbf{z}_t = [N_t,\, E_t]^\top$ is the phone position in local meters.
Weights are normalized to sum to one.

\textbf{Resampling.}
Particle degeneracy is monitored via the effective sample size
$\hat{N}_{\mathrm{eff}} = \bigl(\sum_{p=1}^{P} (w_t^{(p)})^2\bigr)^{-1}$.
When $\hat{N}_{\mathrm{eff}} < P/2$, systematic resampling~\cite{pf} is applied: $P$
new particles are drawn from the current weighted distribution and weights are reset to
$1/P$. Systematic resampling is preferred over multinomial resampling as it reduces
variance while maintaining the same expected particle counts.

\textbf{Estimate.}
After processing all cluster members, the final position estimate is the weighted mean:
\begin{equation}
\hat{N}_C^{\mathrm{PF}} = \sum_{p=1}^{P} w_{n_C}^{(p)}\, x_{N,n_C}^{(p)}, \quad
\hat{E}_C^{\mathrm{PF}} = \sum_{p=1}^{P} w_{n_C}^{(p)}\, x_{E,n_C}^{(p)}
\label{eqn:pf_est}
\end{equation}
converted back to degrees via (\ref{eqn:ekf_back_lat})
and~(\ref{eqn:ekf_back_lon}).

\subsubsection{Post-Clustering Deep Neural Network}
\label{sec:dnn}

Prior work has demonstrated that a GPS baseline position can be refined by a learned
corrector that maps receiver-observable quality features to a position
offset~\cite{siemuri2021, kanhere2022improving}. These approaches operate on raw
per-satellite measurements (pseudorange residuals, line-of-sight vectors, and satellite
geometry angles) produced by a single receiver at a single epoch. In GeoFusion, the
clustering step produces a qualitatively different input: a cluster representative that
already reflects the cooperative fusion of multiple phones across multiple epochs at
the same location event. We therefore propose a \textit{post-clustering deep neural
network} (Post-C DNN) estimator that operates downstream of fusion, taking the cluster
representative and the mean quality features of its members as a fixed-length input,
and regressing a position correction.

\textbf{Input Representation.}
Let $C$ be a cluster of $n_C$ observations. The cluster representative is
$\boldsymbol{\rho}_C = (\phi_C, \lambda_C)$, the centroid or medoid of the reported phone positions within $C$, depending on the clustering algorithm.
The cluster-level quality feature vector is:
\begin{equation}
\bar{\mathbf{x}}_C = \frac{1}{n_C} \sum_{i \in C} \mathbf{x}_i
\label{eqn:dnn_features}
\end{equation}
where each per-observation vector $\mathbf{x}_i$ contains eight epoch-level GNSS
quality observables: $J_{\mathrm{avg}}$, $\mathrm{speedMps}$,
$n_{\mathrm{signals}}$, $\overline{\sigma}_{\mathrm{pr}}$, $h_{\mathrm{DOP}}$,
$v_{\mathrm{DOP}}$, $\bar{\delta}_{\mathrm{iono}}$ (\texttt{avg\_iono}), and
$\bar{\delta}_{\mathrm{tropo}}$ (\texttt{avg\_tropo}). These are exactly the features
available in the GeoFusion dataset and require no raw per-satellite measurements. The
full DNN input is:
\begin{equation}
\mathbf{z}_C = \bigl[\phi_C,\; \lambda_C,\; \bar{\mathbf{x}}_C\bigr] \in \mathbb{R}^{10}
\label{eqn:dnn_input}
\end{equation}

\textbf{Correction Target.}
Rather than predicting absolute coordinates, the DNN predicts a position correction
$(\Delta\phi_C,\, \Delta\lambda_C)$ relative to the cluster representative:
\begin{equation}
\Delta\phi_C = \bar{\phi}_C^{\mathrm{gt}} - \phi_C, \quad
\Delta\lambda_C = \bar{\lambda}_C^{\mathrm{gt}} - \lambda_C
\label{eqn:dnn_delta}
\end{equation}
where $\bar{\phi}_C^{\mathrm{gt}}$ and $\bar{\lambda}_C^{\mathrm{gt}}$ are the mean
ground-truth coordinates of the cluster members. This formulation keeps the learning
target near zero (empirical standard deviation $\approx 5\times10^{-5}$ degrees, or
approximately 5 m), making training well-conditioned.

\textbf{Architecture.}
The Post-C DNN is a multilayer perceptron with three hidden layers of dimensions
$[64, 32, 16]$, each followed by a ReLU activation and dropout at rate $p = 0.1$,
and a linear output layer projecting to $\mathbb{R}^2$. Letting $\hat{\Delta\phi}_C$
and $\hat{\Delta\lambda}_C$ denote the predicted corrections, the training objective
is mean absolute error in meters:
\begin{equation}
\mathcal{L} = \frac{1}{|\mathcal{D}|} \sum_{C \in \mathcal{D}} \Bigl[
\bigl((\hat{\Delta\phi}_C - \Delta\phi_C)\, m_{\mathrm{lat}}\bigr)^2
+ \bigl((\hat{\Delta\lambda}_C - \Delta\lambda_C)\, m_{\mathrm{lon}}\bigr)^2
\Bigr]^{\frac{1}{2}}
\label{eqn:dnn_loss}
\end{equation}
where $\mathcal{D}$ is the training set of clusters and $m_{\mathrm{lat}}$,
$m_{\mathrm{lon}}$ are the WGS84 conversion factors evaluated at $\phi_C$. Both
inputs and targets are independently standardized to zero mean and unit variance using
training-split statistics only.

\textbf{Training Protocol.}
The dataset is partitioned at the drive level: training drives fit the MLP, and
held-out drives provide the early stopping signal. The model is trained with the Adam
optimizer at a learning rate of $10^{-3}$, a batch size of 64 clusters, and early
stopping with patience of 20 epochs over a maximum of 200 epochs. At inference, the
refined position estimate is:
\begin{equation}
\hat{\phi}_C^{\mathrm{DNN}} = \phi_C + \hat{\Delta\phi}_C, \quad
\hat{\lambda}_C^{\mathrm{DNN}} = \lambda_C + \hat{\Delta\lambda}_C
\label{eqn:dnn_prediction}
\end{equation}
\section{Experimental Evaluation}
\label{sec:experiments}

\subsection{Experimental Design}
The clustering stage of GeoFusion is benchmarked against k-means and k-medoids as certain baselines, their uncertainty-aware counterparts UK-means and UK-medoids, and SDSGC, a graph-based algorithm that models uncertainty at the topology level. For the uncertain object algorithms UK-means and UK-medoids, we further test two versions of each model corresponding to the two uncertain object models, the volcano function and the mountain function. Additionally, since the graph algorithm SDSGC does not directly compute a cluster representative, we compute both the centroid and the medoid of the final clusters. Thus, there are a total of eight clustering configurations: the baselines k-means and k-medoids, UK-means (volcano) and UK-means (mountain), UK-medoids (volcano) and UK-medoids (mountain), and SDSGC (centroid) and SDSGC (medoid), all shown in Table~\ref{tab:algorithms}. We run each of the 8 configurations on 11 samples of the data, described in Table~\ref{tab:datasets}, to understand how the performance of the algorithms varies with scale. Therefore, in total, there are 88 clustering experiments, with each configuration running on each sample.

To evaluate the clustering results, we consider three groups of metrics:
\begin{itemize}
    \item \textbf{Cluster quality metrics:}
    \begin{itemize}
        \item \textit{Adjusted Rand Index (ARI):} measures the similarity between the predicted cluster assignments and the ground-truth location event labels, corrected for chance. Higher is better, with 1.0 indicating a perfect match.
        \item \textit{Davies--Bouldin Index (DBI):} measures the average ratio of within-cluster scatter to between-cluster separation. Lower is better, with 0 indicating perfectly compact and well-separated clusters.
        \item \textit{Average correct cluster assignment rate:} the majority-class purity of each predicted cluster with respect to ground-truth labels, averaged across all clusters. Higher is better.
    \end{itemize}
    \item \textbf{Distance error metrics:}
    \begin{itemize}
        \item \textit{Mean absolute error (MAE):} the Euclidean distance between the cluster representative and the ground-truth position of its members, averaged across all clusters.
        \item \textit{North positioning error (N error):} the absolute north component of the displacement vector between the cluster representative and the ground-truth position, averaged across all clusters.
        \item \textit{East positioning error (E error):} the absolute east component of the same displacement vector, averaged across all clusters.
    \end{itemize}
    \item \textbf{Runtime and memory footprint:}
    \begin{itemize}
        \item \textit{Runtime:} wall-clock execution time in seconds measured from the start to the end of the clustering step.
        \item \textit{Peak RAM:} the maximum memory footprint in megabytes recorded during execution using tracemalloc.
    \end{itemize}
\end{itemize}

To manage the total number of experiments, the two top-performing clustering configurations across these metrics are selected for the experiments of the final stage of GeoFusion: location event estimation enhancement. In this stage, the cluster representatives are considered the baseline estimates of the respective location events. We evaluate four post-clustering estimators against this baseline: KF, EKF, PF, and DNN, as shown in Table~\ref{tab:estimators}. Since we select 2 clustering configurations and evaluate their estimation results using 5 different schemes across 11 samples, the total number of experiments for post-clustering location event estimation is 110. For consistency in evaluation, each experiment is assessed using distance error metrics, runtime, and memory footprint.

In total, there are 198 experiments: 88 from the uncertainty representation and clustering stage, and 110 from the post-clustering location event estimation stage. All experiments were conducted in Python on a single-core Linux 6.18.5 containerized environment with 4.19 GB RAM. Our Python repository can be found on GitHub: \url{https://github.com/baahmed/geofusion}. The \texttt{geofusion} package used to construct the pipelines corresponding to the experiments can be found on PyPI: \url{https://pypi.org/project/geofusion/}.

\begin{table}[t]
\centering
\caption{Dataset samples used in the benchmarking study. Each sample is constructed such that the number of ground-truth location events equals $k$ with a total of $n$ observations.}
\label{tab:datasets}
\renewcommand{\arraystretch}{1.15}
\begin{tabular}{cc}
\toprule
Location Events ($k$) & Observations ($n$) \\
\midrule
10  & 573   \\
50  & 1{,}957 \\
100 & 3{,}018 \\
150 & 3{,}762 \\
200 & 4{,}303 \\
250 & 4{,}718 \\
300 & 5{,}054 \\
350 & 5{,}336 \\
400 & 5{,}586 \\
450 & 5{,}803 \\
500 & 6{,}003 \\
\bottomrule
\end{tabular}
\end{table}

\begin{table}[t]
\centering
\caption{Clustering algorithms evaluated in the benchmarking study, grouped by paradigm.}
\label{tab:algorithms}
\renewcommand{\arraystretch}{1.15}
\begin{tabular}{ll}
\toprule
\textbf{Group} & \textbf{Algorithm} \\
\midrule
\multirow{2}{*}{Certain baselines}      & k-means \\
                                         & k-medoids \\
\midrule
\multirow{4}{*}{Uncertain partitional}  & UK-means (volcano) \\
                                         & UK-means (mountain) \\
                                         & UK-medoids (volcano) \\
                                         & UK-medoids (mountain) \\
\midrule
\multirow{2}{*}{Graph-based}            & SDSGC (centroid) \\
                                         & SDSGC (medoid) \\
\bottomrule
\end{tabular}
\end{table}

\begin{table}[t]
\centering
\caption{Post-clustering position estimators evaluated in the benchmarking study.}
\label{tab:estimators}
\renewcommand{\arraystretch}{1.15}
\begin{tabular}{ll}
\toprule
\textbf{Type} & \textbf{Estimator} \\
\midrule
Baseline        & Cluster representative \\
\midrule
\multirow{3}{*}{Filter-based} & Kalman Filter (KF) \\
                               & Extended Kalman Filter (EKF) \\
                               & Particle Filter (PF) \\
\midrule
Learning-based  & Deep Neural Network (DNN) \\
\bottomrule
\end{tabular}
\end{table}

\subsection{Hyperparameter Tuning}
The number of clusters $k$ for partitional algorithms was determined via the elbow method applied to the within-cluster sum of squared distances over a range of candidate values, which consistently recovered the expected number of ground-truth location events across all samples. For SDSGC, hyperparameters were selected via a two-stage search: in the coarse stage, $n_{\mathrm{nbr}}$ and $\gamma$ were swept jointly over a fixed grid with $\eta$ and $\rho$ held constant, and in the fine stage, the remaining parameters were refined per target $k$ by running until the connected components of $W$ reached exactly $k$ for small $k$, retaining the closest-to-$k$ run across multiple restarts for moderate $k$, and applying a single ALM iteration followed by an adaptive $W$-matrix threshold sweep for large $k$.

\subsection{Results}

\subsubsection{Uncertain Representation and Clustering}
\begin{figure*}[t]
    \centering
    \includegraphics[width=\textwidth]{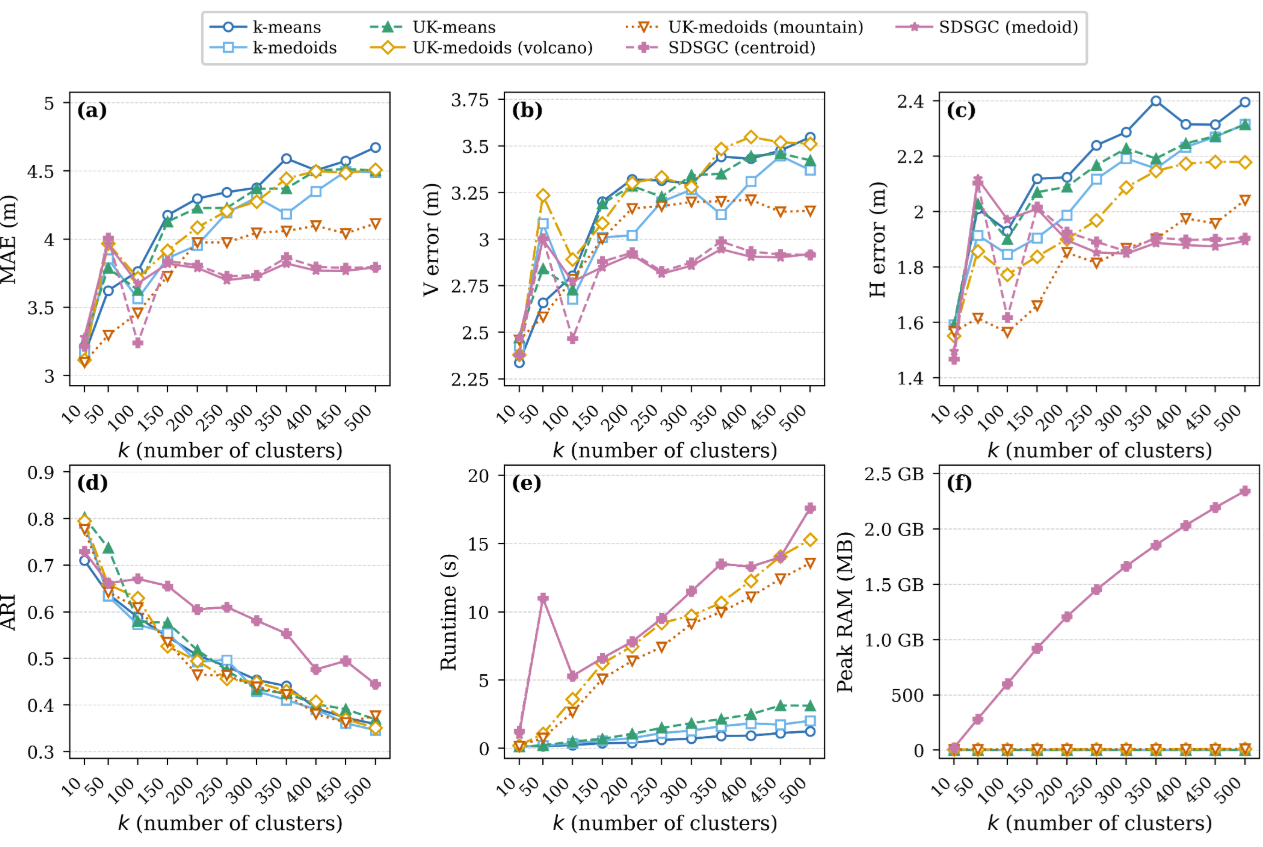}
    \caption{Performance comparison of clustering algorithms across $k \in \{10, 50, 100, 150, 200, 250, 300, 350, 400, 450, 500\}$ on the dataset. Panels show (a) mean absolute error (MAE), (b) north positioning error, (c) east positioning error, (d) Adjusted Rand Index (ARI), (e) runtime, and (f) peak RAM usage. UK-means (volcano) and UK-means (mountain) produce identical results in all experiments, confirming their theoretical equivalence.}
    \label{fig:result_graph}
\end{figure*}

The uncertainty representation and clustering results are shown in Fig.~\ref{fig:result_graph}. Full numerical results for the 88 pre-clustering representation and clustering experiments are reported in Appendix~\ref{appendix:tables}.

\textbf{Cluster Quality at Scale.}
Although both ARI and correct cluster assignment rate degrade as $k$ increases relative to the true number of location events, the absolute values remain strong throughout the experimental range. The two metrics do not always peak for the same algorithm; in roughly half the experiments they are led by different methods, reflecting the fact that cluster purity and structural alignment to the ground truth are related but distinct objectives. The correct assignment rate is highest at small $k$, where the partitioning ratio is close to one, and stabilizes at a consistently high level beyond $k=150$ as the algorithms reach a practical ceiling. DBI follows a complementary trajectory: compact and well-separated clusters are easiest to achieve at small $k$ and become progressively harder as $k$ grows, yet the best-performing algorithms maintain a stable and competitive DBI across the upper range of $k$. Taken together, the two metrics confirm that GeoFusion's clustering quality degrades gracefully with scale rather than collapsing, and that the algorithms retain meaningful discriminative structure throughout the experimental range. Overall, the best-performing clustering algorithm in terms of ARI is UK-means at small $k$ ($k \leq 50$) and SDSGC at all larger values of $k$, where it leads by a consistent margin over all other algorithms.

\textbf{Positioning Error.}
At small $k$, the directional uncertainty model gives UK-medoids (mountain) a sustained advantage on east error, which it holds from $k=50$ through $k=250$ by exploiting the asymmetric north and east scale parameters to place its representative closer to the true location along the east axis. Beyond $k=250$, SDSGC (medoid) takes over as the best-performing algorithm on all three position error metrics simultaneously, a lead it holds through $k=500$ with the sole exception of north error at $k=500$, where the centroid variant leads by a negligible margin. Notably, SDSGC (medoid) errors stabilize markedly from $k=200$ onward: east error remains within a narrow band around $1.87$\,m, north error around $2.90$\,m, and MAE around $3.78$\,m, regardless of further increases in $k$. This plateau indicates that once the graph structure has sufficiently fragmented into the target number of components, the medoid representative converges to a stable position relative to the ground truth, and the additional cluster granularity neither helps nor hurts positional accuracy.

\textbf{Benefits of Uncertainty Awareness.}
Incorporating uncertainty into the clustering objective yields measurable and consistent improvements in position accuracy. UK-medoids (mountain) outperforms k-medoids on MAE in 10 out of 11 experiments, with an average improvement of $0.23$\,m across all eleven values of $k$ (Wilcoxon signed-rank test: $W=65$, $p=0.002$). The benefit is durable across scale: it is present at small $k$, where clusters are large and observations are plentiful, and persists and even grows at large $k$, where tighter clusters make the uncertainty-weighted medoid selection most discriminating. Of the two uncertain object models, the empirically derived mountain function outperforms the geometry-only volcano function, with UK-medoids (mountain) achieving strictly lower MAE than UK-medoids (volcano) across all eleven values of $k$ (Wilcoxon signed-rank test: $W=66$, $p<0.001$), confirming that incorporating additional receiver observables beyond satellite geometry yields a more discriminative uncertainty representation. On cluster purity, however, the picture is more nuanced: UK-medoids leads on correct assignment in only four out of eleven experiments, suggesting that uncertainty awareness improves the position of the representative within the cluster more reliably than it improves the assignment of observations to clusters. UK-means, by contrast, produces only marginal and inconsistent improvements over k-means on MAE, and negligible differences on cluster purity.

\textbf{Runtime and Memory.}
Runtime increases approximately linearly with $k$ for k-means, k-medoids, and UK-means, while UK-medoids and SDSGC exhibit super-linear growth. SDSGC's memory footprint scales quadratically with $n$ due to its $n \times n$ dense affinity matrix $W$, growing from 22\,MB at $n=573$ ($k=10$) to 2,340\,MB at $n=6{,}003$ ($k=500$). The RAM consumed by k-means, k-medoids, and UK-means remains stable across all values of $k$ (1.1--1.6\,MB), since their memory cost depends on the data matrix rather than on $k$. UK-medoids requires more memory due to the pairwise distance computations in the expected-distance metric, growing from 6.3\,MB at $k=10$ to 11.0\,MB at $k=500$.
Full runtime and RAM results are reported in Appendix~\ref{appendix:tables}.

Taken together, these results show that UK-medoids (mountain) is the strongest configuration at small to moderate $k$, leading on position accuracy through $k=150$, while SDSGC (medoid) becomes the dominant configuration from $k=200$ onward, where it leads consistently on all position error metrics. Thus, we use these two configurations as the basis for the last stage of GeoFusion, the post-clustering position estimation, whose results we discuss next.

\subsubsection{Post-Clustering Position Estimation}
The post-clustering position estimation results are shown in Fig.~\ref{fig:result_estimator_graph}. Full numerical results for all 110 experiments are reported in Appendix~\ref{appendix:estimation_tables}.

\begin{figure*}[!t]
    \centering
    \includegraphics[width=\textwidth]{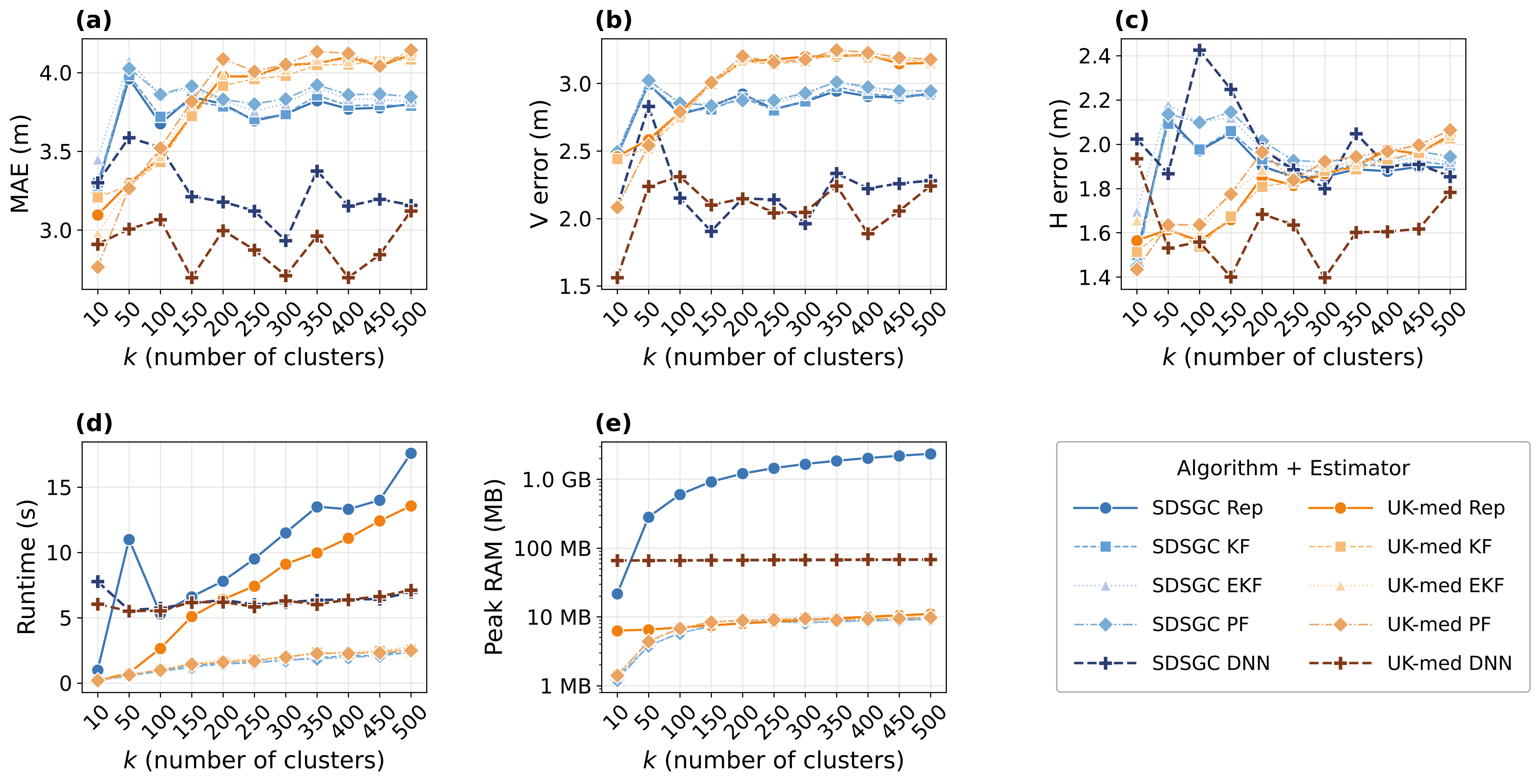}
    \caption{Performance comparison of post-clustering estimators across $k \in \{10, 50, 100, 150, 200, 250, 300, 350, 400, 450, 500\}$ on the dataset. Panels show (a) mean absolute error (MAE), (b) north positioning error, (c) east positioning error, (d) runtime, and (e) peak RAM usage. Results are shown for the baseline cluster representative (Rep) and the four post-clustering estimators applied to each of the two clustering configurations retained from the preceding stage: SDSGC (medoid) and UK-medoids (mountain). Runtime and peak RAM for Rep include the full clustering step; for all other estimators, they cover only the estimation step. DNN results are evaluated on held-out test drives only and are not directly comparable to the representative, KF, EKF, and PF results, which are evaluated on all observations.}
    \label{fig:result_estimator_graph}
\end{figure*}

\textbf{Positioning Error.}
The post-clustering estimators reveal a consistent and interpretable pattern across both clustering algorithms and all values of $k$. For SDSGC clusters, the cluster representative (medoid) is already a strong position estimate: the KF improves on it in only three out of eleven experiments, and even then by negligible margins, while the EKF and PF degrade accuracy in all or nearly all experiments. This suggests that the graph-based medoid captures a near-optimal representative position, and that the sequential filtering assumptions (static target and Gaussian noise) are poorly matched to an already-fused cluster output.

For UK-medoids clusters, the KF provides a small but consistent benefit, improving on the representative in nine out of eleven experiments, while the EKF and PF again offer no reliable gain. The post-clustering DNN, evaluated on held-out test drives, attains lower MAE than the cluster representative attains on the full sample in ten out of eleven experiments for SDSGC and in all eleven for UK-medoids, with average differences of $0.49$\,m and $0.91$\,m respectively. Because the DNN and the representative are evaluated on different sets of observations, this comparison is indicative rather than a matched test. The gap is generally larger at larger $k$, as the training set expands, and UK-medoids consistently provides a better foundation for the DNN than SDSGC, likely because its probabilistic medoid selection produces cluster representatives with more regular residual errors that the network can learn to correct. Overall, UK-medoids (mountain) with the post-clustering DNN achieves a mean MAE of $2.90 \pm 0.15$\,m on held-out test drives across all values of $k$, the lowest error of any configuration evaluated.

\textbf{Runtime and Memory.}
From a computational standpoint, the filter-based estimators are lightweight: KF, EKF, and PF each complete in under three seconds and peak below 10\,MB across all values of $k$, while the DNN requires approximately six to seven seconds and around 67\,MB regardless of $k$, dominated by PyTorch initialization and early stopping rather than by training data volume. Whether clustering or estimation dominates the total cost depends on the clustering algorithm. For SDSGC, clustering dominates memory from $k=50$ onward, with peak RAM growing quadratically with $n$ to 2,340\,MB at $k=500$, and dominates runtime from $k=150$ onward, reaching 17.6\,s at $k=500$. For UK-medoids, clustering remains lightweight in memory, peaking at 11.0\,MB at $k=500$, so the filters add a comparable footprint and the DNN's 67\,MB is the largest memory cost at every $k$; in runtime, UK-medoids clustering grows from under 0.2\,s at $k=10$ to 13.6\,s at $k=500$ and exceeds the DNN's cost from $k=200$ onward.

\subsubsection{Recommended GeoFusion Configuration}
Assuming sufficient training data and no hard constraints on clustering runtime or memory, the recommended GeoFusion configuration is UK-medoids (mountain) with the post-clustering DNN across all values of $k$. UK-medoids provides the stronger clustering foundation for the DNN: its probabilistic medoid selection produces cluster representatives with more regular residual errors that the network can learn to correct, and this configuration achieves the lowest error of all evaluated configurations, a mean MAE of $2.90 \pm 0.15$\,m on held-out test drives. The DNN adds approximately six to seven seconds and 67\,MB regardless of $k$; this is modest in absolute terms, although it exceeds the memory of UK-medoids clustering itself (at most 11.0\,MB) and, for $k \leq 150$, its runtime. Where the DNN cannot be used, the KF provides a lightweight post-clustering step at the cost of the DNN's accuracy gains: UK-medoids (mountain) with the KF is the fallback at $k \leq 150$, and SDSGC (medoid) with the KF is the more accurate fallback from $k \geq 200$ onward. Where clustering runtime or memory is constrained, UK-medoids (mountain) with the KF is preferable at all values of $k$, since SDSGC's peak RAM grows quadratically with $n$ to 2,340\,MB at $k=500$, whereas UK-medoids remains at or below 11.0\,MB.
\section{Conclusion}
\label{sec:conc}

\noindent
In this work, we devised GeoFusion, a framework that fuses observations from multiple smartphones of different models, each with typical positioning errors of several meters, to estimate the position of a location event. On real data, the recommended configuration achieves a mean absolute error of 2.90 $\pm$ 0.15 meters on held-out test drives, comparable to and on average below the best reported single-phone precision of approximately 3 meters \cite{fu2020}. This result shows the potential of determining location from multiple observations by multiple phones rather than from a single phone, whose error can reach up to 20 meters. A key finding is that uncertainty modeling is a primary driver of these precision gains: the empirically derived mountain function outperforms the geometry-only volcano function ($W=66$, $p<0.001$), and uncertain clustering via UK-medoids outperforms its certain counterpart, k-medoids ($W=65$, $p=0.002$). The post-clustering DNN further refines the cluster estimates, yielding the lowest error of all evaluated configurations. The GeoFusion framework was designed to bypass the limitations of other GPS precision improvement methods by integrating into a system with minimal modifications, using naturally occurring data and providing a mechanism for embedding uncertainty. The contributions of this work are: two uncertain object models for GPS observations grounded in satellite geometry and empirical error statistics; a systematic benchmarking study of partitional and graph-based uncertain clustering algorithms across scale; and a post-clustering estimation framework evaluated on real smartphone data.

The GeoFusion framework is made up of three stages: pre-clustering uncertainty representation, clustering observations by location event, and post-clustering location event position estimation. In the first stage, pre-clustering uncertainty representation, we proposed two uncertainty models: the volcano function, based on the average satellite geometry cost of \cite{pdf}, and the empirically validated mountain function, based on sources of error, namely satellite geometry cost, vehicle speed, number of visible signals, and pseudorange uncertainty. In the second stage, clustering, we considered the partitional models k-means and k-medoids and their uncertain counterparts, UK-means and UK-medoids, in addition to SDSGC, a graph-based clustering algorithm. We selected the best-performing configurations from the first two stages and refined their outputs in the final stage, post-clustering location event position estimation, which aims to improve the location estimate relative to the baseline cluster representative using KF, EKF, PF, and DNN.

We recommend applying GeoFusion periodically on the cloud to detect location events and identify their position with high precision as follows. To avoid the complications of extreme observations and high noise, datasets should be cleaned by removing inconsistently logged features, restricting the data to complete cases across modeling features, and screening gross positioning outliers prior to applying the GeoFusion framework. In this study, outliers were identified using the reference positions, which are unavailable in deployment; a deployed system would instead need a reference-free criterion, such as thresholds on receiver quality indicators. The recommended GeoFusion configuration is UK-medoids (mountain) with the post-clustering DNN, which achieves an average MAE of $2.90$\,m on held-out test drives across all evaluated values of $k$, with the DNN adding approximately six to seven seconds and 67\,MB regardless of $k$. Where the DNN cannot be used, UK-medoids (mountain) with the KF at $k \leq 150$ and SDSGC (medoid) with the KF at $k \geq 200$ serve as lightweight post-clustering alternatives; where clustering runtime or memory is constrained, UK-medoids (mountain) with the KF is preferable at all values of $k$.

GeoFusion's experimental evaluation is subject to several limitations that inform the scope of its current claims. The benchmarking study is conducted on samples of up to 6,003 observations across 500 location events drawn from the GSDC dataset, which covers 29 drives in a limited set of environments in the San Francisco Bay Area using 7 distinct phone models. While the progressive sample design provides insight into scaling behavior, the absolute scale remains modest, and broader generalization to larger deployments, more diverse geographies, and a wider range of device types requires further validation. Moreover, the benchmarking samples are drawn from well-observed location events, averaging between 12 and 57 observations per event (Table~\ref{tab:datasets}), whereas the median location event in the raw dataset is observed only once; GeoFusion's performance under such sparse observation, where it reduces to the single-phone fix, requires further validation. At scale, SDSGC's quadratic memory footprint becomes a practical bottleneck, which is why UK-medoids (mountain) with the post-clustering DNN is the primary recommendation despite SDSGC's competitive position error metrics at large $k$. Additionally, all experiments were conducted in a single-core environment, so the reported runtimes do not reflect the potential gains of parallelization. When uncertainty modeling is inaccurate, for instance due to atypical multipath conditions not captured by the modeled features, the uncertain object representation may misweight observations during clustering, potentially degrading the medoid selection. Beyond scale, GeoFusion carries two structural assumptions that bound its applicability. First, the framework operates on two-dimensional latitude/longitude coordinates; altitude is not modeled, and extension to three-dimensional positioning would require modifications to both the uncertain object representation and the clustering step. Second, GeoFusion assumes reliable data transmission from sensors to the cloud, which may not hold in bandwidth-constrained or real-time edge deployment scenarios.

Future work could address these limitations in several directions: extending the uncertain object models to three-dimensional positioning to support altitude-aware applications; infusing multipath errors into the uncertain object representations; validating the framework on larger and more geographically diverse datasets collected across a broader range of GPS devices and sensor configurations to assess generalizability beyond the San Francisco Bay Area; evaluating GeoFusion under sparse and low-density observation conditions, where most location events are observed only a few times; exploring on-device inference or federated aggregation to relax the cloud transmission assumption; and investigating parallelization strategies to reduce the runtime overhead of UK-medoids and SDSGC at scale.

\clearpage
\twocolumn

\bibliographystyle{IEEEtran}
\bibliography{geofusion_ref}

\begin{IEEEbiography}[{\includegraphics[width=1in,height=1.25in,clip,keepaspectratio]{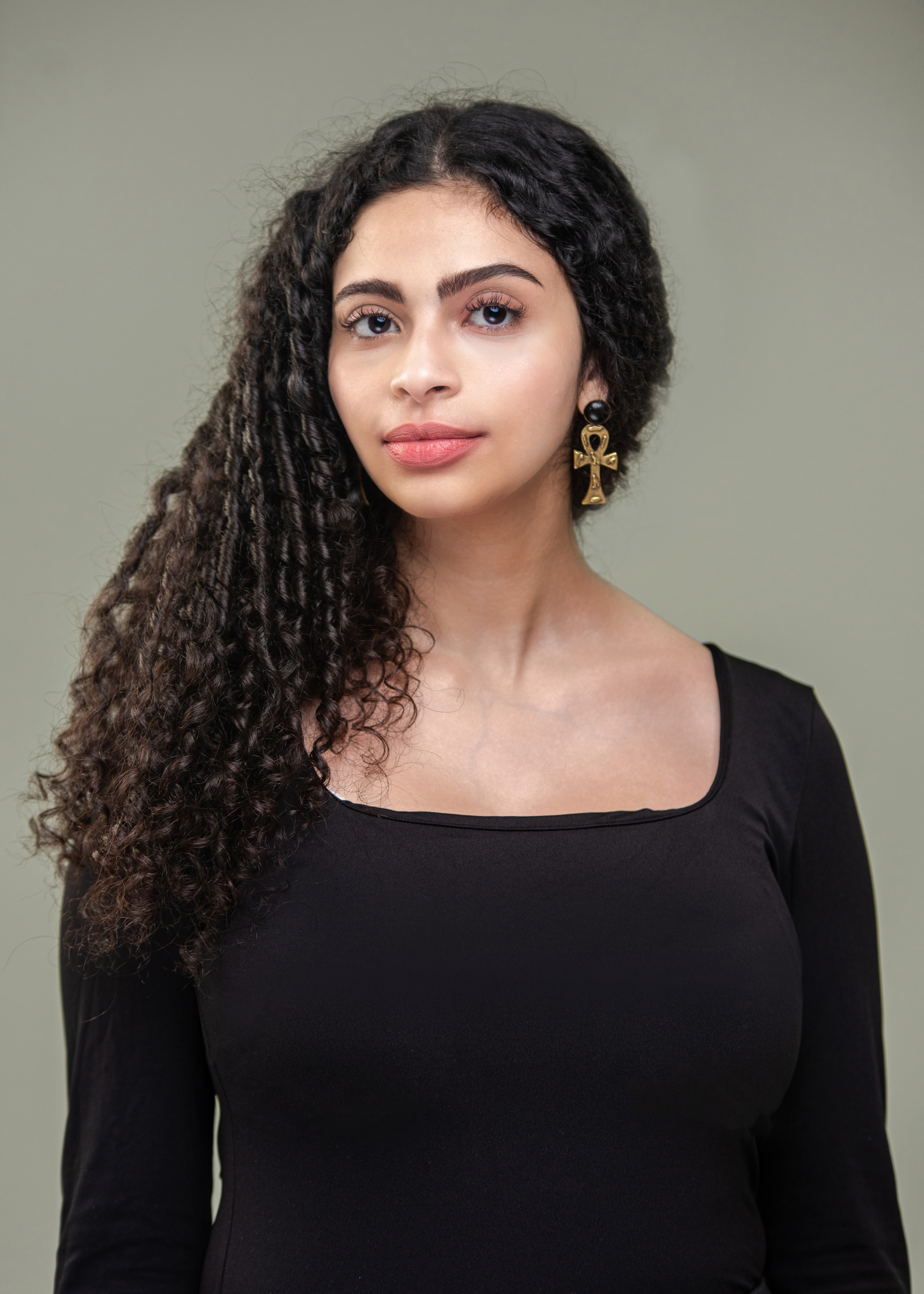}}]{Basant Ahmed Mounir Abdelhamid} is a Data and AI Solutions Consultant for Disaster Risk Reduction at the United Nations Development Programme (UNDP). Previously, Mounir was the Applied Artificial Intelligence Lead at the Central Bank of Egypt. She led her team to build Data and AI solutions across competition law, internal auditing, consumer protection, banking regulations, monetary policy, and sustainability. 

Mounir is deeply committed to mentorship and community. She teaches in MIT's Applied AI and Data Science Program, trained central bankers across Africa in AI through COMESA, and ranks among the top 1\% of data mentors on Topmate. Globally, Mounir is recognized as an active voice in AI. She speaks regularly in international forums, such as Ai4, a global top 10 AI conference, and DSC MENA. She reviews books, most recently Foundations of Agentic AI for Retail. She advises founders worldwide and angel-invested in Mundo AI, supporting its journey from ideation to YC acceptance.

Basant received her B.Sc. in Computer Science and Engineering from the German University in Cairo (GUC). During her time as an undergraduate, she joined the IEEE GUC Student Branch as a hardware specialist and later sought to evangelize data and AI culture. She founded the award-winning data science school with data analytics and machine learning specializations, where she mentored more than 50 students. Promoted to the technical president, Basant led the branch to the \textit{Best Technical Club} award from the council of active working groups. Furthermore, she contributed to the organization of prominent conferences, such as the National Radio Science Conference (NRSC 2020) and the 33rd International Conference on Microelectronics (IEEE-ICM 2021).
\end{IEEEbiography}

\begin{IEEEbiography}[{\includegraphics[width=1in,height=1.25in,clip,keepaspectratio]{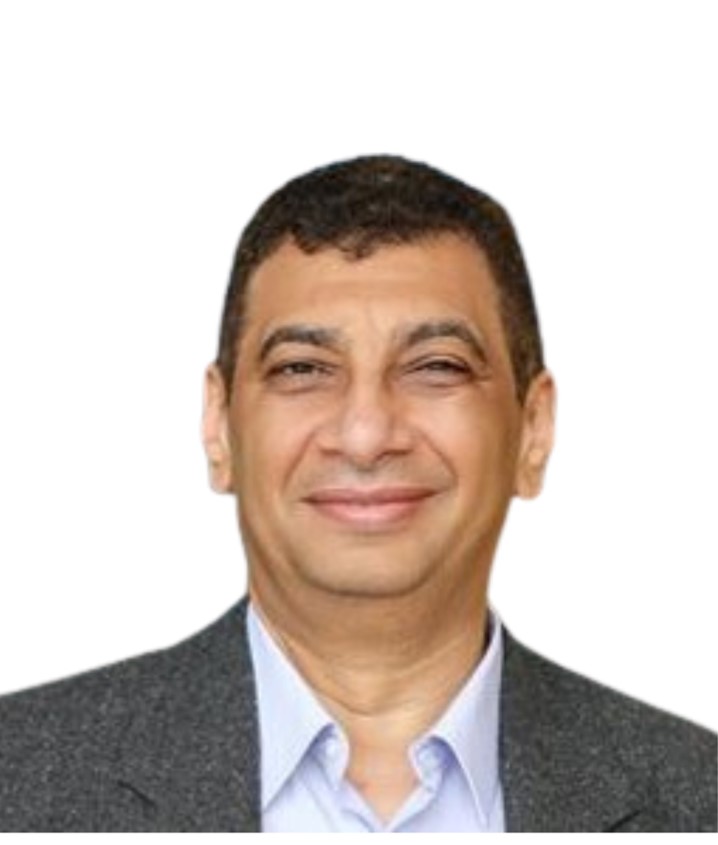}}]{Amr T. Abdel-Hamid}
(Senior Member, IEEE) is a Professor of Innovation and Intelligent Systems at King Abdullah University of Science and Technology (KAUST), Saudi Arabia. He is also affiliated with the German University in Cairo (GUC), where he has served since 2006.

He received the B.Eng. degree in Electronics and Communications Engineering from Cairo University, Egypt, in 1997, and the M.Sc. and Ph.D. degrees in Computer Engineering from Concordia University, Montreal, Canada, in 2001 and 2005, respectively.

Abdel-Hamid's research interests include hardware verification and formal methods, embedded systems architecture, secure and low-power wireless systems, Internet of Things (IoT), and Artificial Intelligence of Things (AIoT). His work focuses on system reliability, correctness, and scalable intelligent architectures that integrate AI with distributed embedded systems. He has authored more than 50 refereed journal and conference publications and has supervised over 250 undergraduate and graduate research projects.

In addition to his research contributions, Amr has played a leading role in innovation-driven engineering education, developing studio-based system design curricula that integrate hardware, software, and AI. His work bridges formal system correctness with practical deployment of intelligent IoT systems in industrial environments. He has led interdisciplinary research and applied innovation initiatives across academia and industry in Europe, North America, and the Middle East. In addition to his academic contributions, he founded SSTM Egypt in 2016, where he leads applied research and development in IoT and AI-enabled systems, bridging academic research with industrial implementation.
\end{IEEEbiography}

% ── Appendices (IEEE order: after the Conclusion, before the References) ──
\clearpage
\appendices
\onecolumn

% Appendix A: heading, label and intro are inside geofusion_tables_A
% ==================================================================
% Appendix A: clustering benchmark (follows the existing \appendices command)
% Requires: booktabs, float
% ==================================================================
\section{Detailed Clustering Benchmark Results}
\label{appendix:tables}

Tables~\ref{tab:mae}--\ref{tab:ram} report full numerical results for all algorithms across $k \in \{10, 50, 100, 150, 200, 250, 300, 350, 400, 450, 500\}$. Algorithms are grouped into certain baselines (k-means, k-medoids), uncertain partitional methods (UK-means and UK-medoids under both uncertainty models), and the graph-based method (SDSGC). Bold denotes the best value per column; tied values are all shown in bold.

\begin{table}[H]
\centering
\caption{Mean absolute error (MAE, m) for each algorithm across all evaluated values of $k$. Bold denotes the best (lowest) value per column.}
\label{tab:mae}
\renewcommand{\arraystretch}{1.15}
\small
\begin{tabular*}{\textwidth}{@{\extracolsep{\fill}}lrrrrrrrrrrr}
\toprule
\textbf{Algorithm} & \textbf{10} & \textbf{50} & \textbf{100} & \textbf{150} & \textbf{200} & \textbf{250} & \textbf{300} & \textbf{350} & \textbf{400} & \textbf{450} & \textbf{500} \\
\midrule
\multicolumn{12}{l}{\textit{Certain baselines}} \\
k-means & 3.15 & 3.62 & 3.76 & 4.17 & 4.30 & 4.34 & 4.38 & 4.59 & 4.50 & 4.57 & 4.67 \\
k-medoids & 3.17 & 3.85 & 3.56 & 3.86 & 3.95 & 4.19 & 4.30 & 4.18 & 4.35 & 4.50 & 4.49 \\
\midrule
\multicolumn{12}{l}{\textit{Uncertain partitional}} \\
UK-means (volcano) & 3.25 & 3.79 & 3.63 & 4.13 & 4.23 & 4.23 & 4.37 & 4.37 & 4.50 & 4.51 & 4.50 \\
UK-means (mountain) & 3.25 & 3.79 & 3.63 & 4.13 & 4.23 & 4.23 & 4.37 & 4.37 & 4.50 & 4.51 & 4.50 \\
UK-medoids (volcano) & 3.12 & 3.97 & 3.71 & 3.92 & 4.08 & 4.21 & 4.27 & 4.44 & 4.50 & 4.49 & 4.51 \\
UK-medoids (mountain) & \textbf{3.10} & \textbf{3.30} & 3.46 & \textbf{3.73} & 3.97 & 3.98 & 4.05 & 4.06 & 4.10 & 4.04 & 4.11 \\
\midrule
\multicolumn{12}{l}{\textit{Graph-based}} \\
SDSGC (centroid) & 3.22 & 4.01 & \textbf{3.24} & 3.84 & 3.81 & 3.73 & 3.74 & 3.86 & 3.80 & 3.79 & 3.80 \\
SDSGC (medoid) & 3.28 & 3.96 & 3.67 & 3.82 & \textbf{3.79} & \textbf{3.70} & \textbf{3.72} & \textbf{3.82} & \textbf{3.77} & \textbf{3.77} & \textbf{3.79} \\
\bottomrule
\end{tabular*}
\end{table}

\begin{table}[H]
\centering
\caption{North positioning error (N error, m) for each algorithm across all evaluated values of $k$. Bold denotes the best (lowest) value per column.}
\label{tab:v_error}
\renewcommand{\arraystretch}{1.15}
\small
\begin{tabular*}{\textwidth}{@{\extracolsep{\fill}}lrrrrrrrrrrr}
\toprule
\textbf{Algorithm} & \textbf{10} & \textbf{50} & \textbf{100} & \textbf{150} & \textbf{200} & \textbf{250} & \textbf{300} & \textbf{350} & \textbf{400} & \textbf{450} & \textbf{500} \\
\midrule
\multicolumn{12}{l}{\textit{Certain baselines}} \\
k-means & \textbf{2.34} & 2.66 & 2.80 & 3.20 & 3.32 & 3.31 & 3.30 & 3.44 & 3.43 & 3.48 & 3.55 \\
k-medoids & 2.43 & 3.08 & 2.68 & 3.01 & 3.02 & 3.20 & 3.27 & 3.13 & 3.31 & 3.45 & 3.37 \\
\midrule
\multicolumn{12}{l}{\textit{Uncertain partitional}} \\
UK-means (volcano) & 2.47 & 2.84 & 2.73 & 3.19 & 3.29 & 3.23 & 3.34 & 3.35 & 3.45 & 3.46 & 3.42 \\
UK-means (mountain) & 2.47 & 2.84 & 2.73 & 3.19 & 3.29 & 3.23 & 3.34 & 3.35 & 3.45 & 3.46 & 3.42 \\
UK-medoids (volcano) & 2.38 & 3.23 & 2.89 & 3.09 & 3.30 & 3.33 & 3.28 & 3.48 & 3.55 & 3.52 & 3.51 \\
UK-medoids (mountain) & 2.46 & \textbf{2.58} & 2.79 & 3.00 & 3.16 & 3.18 & 3.20 & 3.20 & 3.21 & 3.15 & 3.15 \\
\midrule
\multicolumn{12}{l}{\textit{Graph-based}} \\
SDSGC (centroid) & 2.38 & 3.00 & \textbf{2.47} & 2.88 & 2.93 & 2.83 & 2.87 & 2.99 & 2.93 & 2.92 & \textbf{2.92} \\
SDSGC (medoid) & 2.47 & 3.00 & 2.77 & \textbf{2.85} & \textbf{2.92} & \textbf{2.81} & \textbf{2.86} & \textbf{2.95} & \textbf{2.91} & \textbf{2.90} & \textbf{2.92} \\
\bottomrule
\end{tabular*}
\end{table}

\begin{table}[H]
\centering
\caption{East positioning error (E error, m) for each algorithm across all evaluated values of $k$. Bold denotes the best (lowest) value per column.}
\label{tab:h_error}
\renewcommand{\arraystretch}{1.15}
\small
\begin{tabular*}{\textwidth}{@{\extracolsep{\fill}}lrrrrrrrrrrr}
\toprule
\textbf{Algorithm} & \textbf{10} & \textbf{50} & \textbf{100} & \textbf{150} & \textbf{200} & \textbf{250} & \textbf{300} & \textbf{350} & \textbf{400} & \textbf{450} & \textbf{500} \\
\midrule
\multicolumn{12}{l}{\textit{Certain baselines}} \\
k-means & 1.59 & 2.01 & 1.93 & 2.12 & 2.12 & 2.24 & 2.29 & 2.40 & 2.31 & 2.31 & 2.39 \\
k-medoids & 1.59 & 1.91 & 1.84 & 1.90 & 1.99 & 2.12 & 2.19 & 2.15 & 2.23 & 2.27 & 2.32 \\
\midrule
\multicolumn{12}{l}{\textit{Uncertain partitional}} \\
UK-means (volcano) & 1.59 & 2.03 & 1.90 & 2.07 & 2.09 & 2.17 & 2.23 & 2.19 & 2.25 & 2.27 & 2.31 \\
UK-means (mountain) & 1.59 & 2.03 & 1.90 & 2.07 & 2.09 & 2.17 & 2.23 & 2.19 & 2.25 & 2.27 & 2.31 \\
UK-medoids (volcano) & 1.55 & 1.86 & 1.77 & 1.84 & 1.90 & 1.97 & 2.09 & 2.15 & 2.17 & 2.18 & 2.18 \\
UK-medoids (mountain) & 1.57 & \textbf{1.61} & \textbf{1.56} & \textbf{1.66} & \textbf{1.85} & \textbf{1.81} & 1.87 & 1.90 & 1.97 & 1.96 & 2.04 \\
\midrule
\multicolumn{12}{l}{\textit{Graph-based}} \\
SDSGC (centroid) & \textbf{1.47} & 2.10 & 1.62 & 2.02 & 1.93 & 1.89 & 1.86 & 1.91 & 1.90 & 1.90 & 1.90 \\
SDSGC (medoid) & 1.50 & 2.12 & 1.97 & 2.01 & 1.90 & 1.85 & \textbf{1.85} & \textbf{1.89} & \textbf{1.88} & \textbf{1.87} & \textbf{1.89} \\
\bottomrule
\end{tabular*}
\end{table}

\begin{table}[H]
\centering
\caption{Adjusted Rand Index (ARI) for each algorithm across all evaluated values of $k$. Bold denotes the best (highest) value per column.}
\label{tab:ari}
\renewcommand{\arraystretch}{1.15}
\small
\begin{tabular*}{\textwidth}{@{\extracolsep{\fill}}lrrrrrrrrrrr}
\toprule
\textbf{Algorithm} & \textbf{10} & \textbf{50} & \textbf{100} & \textbf{150} & \textbf{200} & \textbf{250} & \textbf{300} & \textbf{350} & \textbf{400} & \textbf{450} & \textbf{500} \\
\midrule
\multicolumn{12}{l}{\textit{Certain baselines}} \\
k-means & 0.710 & 0.638 & 0.588 & 0.547 & 0.507 & 0.482 & 0.454 & 0.441 & 0.393 & 0.372 & 0.359 \\
k-medoids & 0.791 & 0.634 & 0.572 & 0.554 & 0.493 & 0.497 & 0.429 & 0.411 & 0.390 & 0.361 & 0.346 \\
\midrule
\multicolumn{12}{l}{\textit{Uncertain partitional}} \\
UK-means (volcano) & \textbf{0.803} & \textbf{0.737} & 0.579 & 0.576 & 0.518 & 0.475 & 0.433 & 0.425 & 0.402 & 0.391 & 0.369 \\
UK-means (mountain) & \textbf{0.803} & \textbf{0.737} & 0.579 & 0.576 & 0.518 & 0.475 & 0.433 & 0.425 & 0.402 & 0.391 & 0.369 \\
UK-medoids (volcano) & 0.795 & 0.659 & 0.630 & 0.526 & 0.494 & 0.456 & 0.448 & 0.430 & 0.407 & 0.369 & 0.351 \\
UK-medoids (mountain) & 0.777 & 0.642 & 0.609 & 0.535 & 0.465 & 0.463 & 0.439 & 0.423 & 0.380 & 0.362 & 0.377 \\
\midrule
\multicolumn{12}{l}{\textit{Graph-based}} \\
SDSGC (centroid) & 0.730 & 0.661 & \textbf{0.671} & \textbf{0.656} & \textbf{0.605} & \textbf{0.610} & \textbf{0.581} & \textbf{0.553} & \textbf{0.476} & \textbf{0.495} & \textbf{0.445} \\
SDSGC (medoid) & 0.730 & 0.661 & \textbf{0.671} & \textbf{0.656} & \textbf{0.605} & \textbf{0.610} & \textbf{0.581} & \textbf{0.553} & \textbf{0.476} & \textbf{0.495} & \textbf{0.445} \\
\bottomrule
\end{tabular*}
\end{table}

\begin{table}[H]
\centering
\caption{Davies--Bouldin Index (DBI) for each algorithm across all evaluated values of $k$. Bold denotes the best (lowest) value per column.}
\label{tab:dbi}
\renewcommand{\arraystretch}{1.15}
\small
\begin{tabular*}{\textwidth}{@{\extracolsep{\fill}}lrrrrrrrrrrr}
\toprule
\textbf{Algorithm} & \textbf{10} & \textbf{50} & \textbf{100} & \textbf{150} & \textbf{200} & \textbf{250} & \textbf{300} & \textbf{350} & \textbf{400} & \textbf{450} & \textbf{500} \\
\midrule
\multicolumn{12}{l}{\textit{Certain baselines}} \\
k-means & 0.596 & 0.580 & \textbf{0.655} & 0.710 & 0.741 & 0.716 & \textbf{0.722} & \textbf{0.711} & 0.710 & \textbf{0.701} & 0.695 \\
k-medoids & 0.555 & 0.651 & 0.668 & 0.720 & 0.755 & 0.746 & 0.731 & 0.753 & 0.736 & 0.715 & \textbf{0.687} \\
\midrule
\multicolumn{12}{l}{\textit{Uncertain partitional}} \\
UK-means (volcano) & 0.567 & 0.627 & 0.669 & \textbf{0.699} & \textbf{0.715} & \textbf{0.715} & 0.725 & 0.720 & \textbf{0.700} & 0.705 & 0.698 \\
UK-means (mountain) & 0.567 & 0.627 & 0.669 & \textbf{0.699} & \textbf{0.715} & \textbf{0.715} & 0.725 & 0.720 & \textbf{0.700} & 0.705 & 0.698 \\
UK-medoids (volcano) & 0.596 & 0.701 & 0.715 & 0.822 & 0.864 & 0.857 & 0.833 & 0.831 & 0.853 & 0.846 & 0.807 \\
UK-medoids (mountain) & 0.609 & 0.803 & 0.921 & 1.072 & 1.153 & 1.134 & 1.318 & 1.234 & 1.449 & 1.377 & 1.389 \\
\midrule
\multicolumn{12}{l}{\textit{Graph-based}} \\
SDSGC (centroid) & \textbf{0.411} & \textbf{0.478} & 0.681 & 0.773 & 0.874 & 0.902 & 0.898 & 0.834 & 0.886 & 0.864 & 0.857 \\
SDSGC (medoid) & \textbf{0.411} & \textbf{0.478} & 0.681 & 0.773 & 0.874 & 0.902 & 0.898 & 0.834 & 0.886 & 0.864 & 0.857 \\
\bottomrule
\end{tabular*}
\end{table}

\begin{table}[H]
\centering
\caption{Average correct cluster assignment rate (majority-class purity per cluster, \%) for each algorithm across all evaluated values of $k$. Bold denotes the best (highest) value per column.}
\label{tab:correct}
\renewcommand{\arraystretch}{1.15}
\small
\begin{tabular*}{\textwidth}{@{\extracolsep{\fill}}lrrrrrrrrrrr}
\toprule
\textbf{Algorithm} & \textbf{10} & \textbf{50} & \textbf{100} & \textbf{150} & \textbf{200} & \textbf{250} & \textbf{300} & \textbf{350} & \textbf{400} & \textbf{450} & \textbf{500} \\
\midrule
\multicolumn{12}{l}{\textit{Certain baselines}} \\
k-means & 91.2 & 93.2 & 90.1 & 86.6 & 85.3 & 86.6 & 85.2 & \textbf{87.9} & 86.3 & 87.1 & 87.2 \\
k-medoids & 93.2 & 91.0 & 89.9 & 87.1 & 85.8 & 86.2 & 86.1 & 86.3 & 86.1 & \textbf{87.5} & 86.6 \\
\midrule
\multicolumn{12}{l}{\textit{Uncertain partitional}} \\
UK-means (volcano) & 93.3 & \textbf{94.6} & 90.1 & 87.3 & \textbf{86.3} & 86.1 & 85.9 & 86.8 & 86.5 & 86.7 & 87.5 \\
UK-means (mountain) & 93.3 & \textbf{94.6} & 90.1 & 87.3 & \textbf{86.3} & 86.1 & 85.9 & 86.8 & 86.5 & 86.7 & 87.5 \\
UK-medoids (volcano) & \textbf{93.4} & 93.1 & 90.8 & 87.3 & 85.7 & \textbf{87.7} & 85.9 & 86.7 & 86.2 & 87.2 & 87.1 \\
UK-medoids (mountain) & 93.1 & 92.5 & 89.8 & 86.3 & 85.3 & 86.4 & 86.7 & 86.4 & 85.4 & 87.1 & 86.0 \\
\midrule
\multicolumn{12}{l}{\textit{Graph-based}} \\
SDSGC (centroid) & 92.0 & 90.9 & \textbf{91.4} & \textbf{87.9} & \textbf{86.3} & 87.4 & \textbf{87.0} & 87.1 & \textbf{86.7} & 87.2 & \textbf{88.0} \\
SDSGC (medoid) & 92.0 & 90.9 & \textbf{91.4} & \textbf{87.9} & \textbf{86.3} & 87.4 & \textbf{87.0} & 87.1 & \textbf{86.7} & 87.2 & \textbf{88.0} \\
\bottomrule
\end{tabular*}
\end{table}

\begin{table}[H]
\centering
\caption{Wall-clock runtime (s) for each algorithm across all evaluated values of $k$. Bold denotes the best (lowest) value per column. UK-means (volcano) and UK-means (mountain) follow identical computation paths; their runtimes differ by at most 0.11\,s.}
\label{tab:time}
\renewcommand{\arraystretch}{1.15}
\small
\begin{tabular*}{\textwidth}{@{\extracolsep{\fill}}lrrrrrrrrrrr}
\toprule
\textbf{Algorithm} & \textbf{10} & \textbf{50} & \textbf{100} & \textbf{150} & \textbf{200} & \textbf{250} & \textbf{300} & \textbf{350} & \textbf{400} & \textbf{450} & \textbf{500} \\
\midrule
\multicolumn{12}{l}{\textit{Certain baselines}} \\
k-means & 0.15 & \textbf{0.15} & \textbf{0.23} & \textbf{0.37} & \textbf{0.40} & \textbf{0.61} & \textbf{0.70} & \textbf{0.90} & \textbf{0.92} & \textbf{1.11} & \textbf{1.24} \\
k-medoids & \textbf{0.10} & 0.19 & 0.36 & 0.56 & 0.74 & 1.11 & 1.29 & 1.61 & 1.82 & 1.73 & 2.01 \\
\midrule
\multicolumn{12}{l}{\textit{Uncertain partitional}} \\
UK-means (volcano) & 0.11 & 0.21 & 0.49 & 0.69 & 1.05 & 1.49 & 1.84 & 2.13 & 2.49 & 3.13 & 3.12 \\
UK-means (mountain) & \textbf{0.10} & 0.21 & 0.48 & 0.71 & 1.07 & 1.38 & 1.91 & 2.19 & 2.51 & 3.16 & 3.12 \\
UK-medoids (volcano) & 0.19 & 1.05 & 3.60 & 6.21 & 7.46 & 9.18 & 9.71 & 10.65 & 12.27 & 14.05 & 15.28 \\
UK-medoids (mountain) & 0.16 & 0.76 & 2.64 & 5.08 & 6.39 & 7.42 & 9.11 & 9.97 & 11.09 & 12.41 & 13.56 \\
\midrule
\multicolumn{12}{l}{\textit{Graph-based}} \\
SDSGC (centroid) & 1.25 & 11.00 & 5.30 & 6.60 & 7.80 & 9.50 & 11.50 & 13.50 & 13.30 & 14.00 & 17.60 \\
SDSGC (medoid) & 0.99 & 11.00 & 5.30 & 6.60 & 7.80 & 9.50 & 11.50 & 13.50 & 13.30 & 14.00 & 17.60 \\
\bottomrule
\end{tabular*}
\end{table}

\begin{table}[H]
\centering
\caption{Peak RAM (MB) for each algorithm across all evaluated values of $k$. Bold denotes the best (lowest) value per column. SDSGC allocates a dense $n \times n$ affinity matrix, producing quadratic memory growth with $n$.}
\label{tab:ram}
\renewcommand{\arraystretch}{1.15}
\small
\begin{tabular*}{\textwidth}{@{\extracolsep{\fill}}lrrrrrrrrrrr}
\toprule
\textbf{Algorithm} & \textbf{10} & \textbf{50} & \textbf{100} & \textbf{150} & \textbf{200} & \textbf{250} & \textbf{300} & \textbf{350} & \textbf{400} & \textbf{450} & \textbf{500} \\
\midrule
\multicolumn{12}{l}{\textit{Certain baselines}} \\
k-means & 1.2 & 1.2 & 1.3 & 1.4 & 1.5 & 1.5 & 1.5 & 1.5 & 1.5 & 1.5 & 1.5 \\
k-medoids & \textbf{1.1} & \textbf{1.1} & \textbf{1.2} & \textbf{1.3} & \textbf{1.4} & \textbf{1.4} & \textbf{1.4} & \textbf{1.4} & \textbf{1.4} & \textbf{1.4} & \textbf{1.4} \\
\midrule
\multicolumn{12}{l}{\textit{Uncertain partitional}} \\
UK-means (volcano) & 1.2 & 1.3 & 1.4 & 1.5 & 1.6 & 1.6 & 1.6 & 1.6 & 1.6 & 1.6 & 1.6 \\
UK-means (mountain) & 1.2 & 1.3 & 1.4 & 1.5 & 1.6 & 1.6 & 1.6 & 1.6 & 1.6 & 1.6 & 1.6 \\
UK-medoids (volcano) & 6.3 & 6.5 & 7.0 & 7.5 & 8.0 & 8.5 & 9.0 & 9.5 & 10.0 & 10.5 & 11.0 \\
UK-medoids (mountain) & 6.3 & 6.5 & 7.0 & 7.5 & 8.0 & 8.5 & 9.0 & 9.5 & 10.0 & 10.5 & 11.0 \\
\midrule
\multicolumn{12}{l}{\textit{Graph-based}} \\
SDSGC (centroid) & 21.6 & 281.5 & 599.0 & 921.0 & 1{,}205.0 & 1{,}449.0 & 1{,}662.0 & 1{,}853.0 & 2{,}030.0 & 2{,}191.0 & 2{,}340.0 \\
SDSGC (medoid) & 21.6 & 281.5 & 599.0 & 921.0 & 1{,}205.0 & 1{,}449.0 & 1{,}662.0 & 1{,}853.0 & 2{,}030.0 & 2{,}191.0 & 2{,}340.0 \\
\bottomrule
\end{tabular*}
\end{table}

\clearpage

% Appendix B: heading, label and intro are inside geofusion_tables_B
% ==================================================================
% Appendix B: post-clustering estimation
% ==================================================================
\section{Detailed Post-Clustering Estimation Results}
\label{appendix:estimation_tables}

Tables~\ref{tab:est_mae}--\ref{tab:est_ram} report full numerical results for the post-clustering estimation experiments across all evaluated values of $k$. Each table reports one metric for the cluster representative (Rep) and the four post-clustering estimators, applied to each of the two clustering configurations retained from the clustering phase: SDSGC (medoid) and UK-medoids (mountain). Bold denotes the best value per column across all estimators; tied values are all shown in bold. In Tables~\ref{tab:est_rt} and~\ref{tab:est_ram}, Rep is excluded from this comparison because its runtime and peak RAM include the full clustering step. DNN results ($\dagger$) are evaluated on held-out test drives only.

\begin{table}[H]
\centering
\caption{Mean absolute error (MAE, m) of the cluster representative and post-clustering estimators across all evaluated values of $k$. Bold denotes the best (lowest) value per column.}
\label{tab:est_mae}
\renewcommand{\arraystretch}{1.15}
\small
\begin{tabular*}{\textwidth}{@{\extracolsep{\fill}}lrrrrrrrrrrr}
\toprule
\textbf{Estimator} & \textbf{10} & \textbf{50} & \textbf{100} & \textbf{150} & \textbf{200} & \textbf{250} & \textbf{300} & \textbf{350} & \textbf{400} & \textbf{450} & \textbf{500} \\
\midrule
\multicolumn{12}{l}{\textit{SDSGC (medoid)}} \\
Rep & 3.28 & 3.96 & 3.68 & 3.85 & 3.80 & 3.69 & 3.74 & 3.82 & 3.77 & 3.78 & 3.80 \\
KF & 3.30 & 3.98 & 3.72 & 3.82 & 3.79 & 3.71 & 3.74 & 3.86 & 3.79 & 3.79 & 3.79 \\
EKF & 3.45 & 4.08 & 3.86 & 3.90 & 3.82 & 3.76 & 3.80 & 3.90 & 3.83 & 3.83 & 3.81 \\
PF & 3.24 & 4.03 & 3.86 & 3.92 & 3.83 & 3.80 & 3.83 & 3.92 & 3.86 & 3.86 & 3.85 \\
DNN$^\dagger$ & 3.30 & 3.59 & 3.53 & 3.21 & 3.18 & 3.12 & 2.93 & 3.38 & 3.15 & 3.20 & 3.16 \\
\midrule
\multicolumn{12}{l}{\textit{UK-medoids (mountain)}} \\
Rep & 3.10 & 3.30 & 3.46 & 3.73 & 3.98 & 3.98 & 4.05 & 4.06 & 4.10 & 4.04 & 4.11 \\
KF & 3.21 & 3.28 & 3.43 & 3.72 & 3.92 & 3.96 & 3.98 & 4.05 & 4.05 & 4.07 & 4.08 \\
EKF & 2.98 & 3.28 & 3.47 & 3.80 & 3.98 & 3.99 & 4.02 & 4.08 & 4.07 & 4.09 & 4.11 \\
PF & \textbf{2.77} & 3.27 & 3.52 & 3.82 & 4.09 & 4.01 & 4.05 & 4.13 & 4.12 & 4.04 & 4.14 \\
DNN$^\dagger$ & 2.91 & \textbf{3.01} & \textbf{3.07} & \textbf{2.70} & \textbf{3.00} & \textbf{2.87} & \textbf{2.71} & \textbf{2.96} & \textbf{2.70} & \textbf{2.84} & \textbf{3.12} \\
\bottomrule
\end{tabular*}
\end{table}

\begin{table}[H]
\centering
\caption{North positioning error (N error, m) of the cluster representative and post-clustering estimators across all evaluated values of $k$. Bold denotes the best (lowest) value per column.}
\label{tab:est_v}
\renewcommand{\arraystretch}{1.15}
\small
\begin{tabular*}{\textwidth}{@{\extracolsep{\fill}}lrrrrrrrrrrr}
\toprule
\textbf{Estimator} & \textbf{10} & \textbf{50} & \textbf{100} & \textbf{150} & \textbf{200} & \textbf{250} & \textbf{300} & \textbf{350} & \textbf{400} & \textbf{450} & \textbf{500} \\
\midrule
\multicolumn{12}{l}{\textit{SDSGC (medoid)}} \\
Rep & 2.47 & 3.00 & 2.77 & 2.83 & 2.92 & 2.81 & 2.87 & 2.95 & 2.91 & 2.90 & 2.93 \\
KF & 2.45 & 2.99 & 2.80 & 2.81 & 2.90 & 2.80 & 2.87 & 2.98 & 2.92 & 2.91 & 2.91 \\
EKF & 2.48 & 3.02 & 2.85 & 2.85 & 2.90 & 2.85 & 2.92 & 3.01 & 2.96 & 2.93 & 2.92 \\
PF & 2.49 & 3.02 & 2.86 & 2.84 & 2.87 & 2.88 & 2.93 & 3.01 & 2.97 & 2.95 & 2.94 \\
DNN$^\dagger$ & 2.09 & 2.83 & \textbf{2.15} & \textbf{1.91} & \textbf{2.15} & 2.14 & \textbf{1.96} & 2.33 & 2.22 & 2.26 & 2.28 \\
\midrule
\multicolumn{12}{l}{\textit{UK-medoids (mountain)}} \\
Rep & 2.46 & 2.58 & 2.79 & 3.00 & 3.16 & 3.18 & 3.20 & 3.20 & 3.21 & 3.15 & 3.15 \\
KF & 2.44 & 2.53 & 2.75 & 3.00 & 3.16 & 3.14 & 3.17 & 3.22 & 3.19 & 3.18 & 3.15 \\
EKF & 2.09 & 2.52 & 2.75 & 3.01 & 3.17 & 3.15 & 3.18 & 3.24 & 3.19 & 3.19 & 3.16 \\
PF & 2.08 & 2.54 & 2.79 & 3.01 & 3.20 & 3.16 & 3.18 & 3.25 & 3.23 & 3.19 & 3.18 \\
DNN$^\dagger$ & \textbf{1.56} & \textbf{2.24} & 2.31 & 2.10 & \textbf{2.15} & \textbf{2.04} & 2.05 & \textbf{2.24} & \textbf{1.89} & \textbf{2.06} & \textbf{2.24} \\
\bottomrule
\end{tabular*}
\end{table}

\begin{table}[H]
\centering
\caption{East positioning error (E error, m) of the cluster representative and post-clustering estimators across all evaluated values of $k$. Bold denotes the best (lowest) value per column.}
\label{tab:est_h}
\renewcommand{\arraystretch}{1.15}
\small
\begin{tabular*}{\textwidth}{@{\extracolsep{\fill}}lrrrrrrrrrrr}
\toprule
\textbf{Estimator} & \textbf{10} & \textbf{50} & \textbf{100} & \textbf{150} & \textbf{200} & \textbf{250} & \textbf{300} & \textbf{350} & \textbf{400} & \textbf{450} & \textbf{500} \\
\midrule
\multicolumn{12}{l}{\textit{SDSGC (medoid)}} \\
Rep & 1.50 & 2.12 & 1.97 & 2.05 & 1.90 & 1.85 & 1.85 & 1.89 & 1.88 & 1.90 & 1.89 \\
KF & 1.50 & 2.09 & 1.98 & 2.06 & 1.93 & 1.89 & 1.87 & 1.91 & 1.90 & 1.92 & 1.91 \\
EKF & 1.70 & 2.18 & 2.10 & 2.12 & 1.97 & 1.91 & 1.89 & 1.93 & 1.92 & 1.94 & 1.92 \\
PF & 1.45 & 2.14 & 2.10 & 2.15 & 2.02 & 1.93 & 1.92 & 1.94 & 1.92 & 1.97 & 1.94 \\
DNN$^\dagger$ & 2.02 & 1.87 & 2.43 & 2.25 & 1.98 & 1.89 & 1.80 & 2.05 & 1.90 & 1.91 & 1.85 \\
\midrule
\multicolumn{12}{l}{\textit{UK-medoids (mountain)}} \\
Rep & 1.57 & 1.61 & 1.56 & 1.66 & 1.85 & 1.81 & 1.87 & 1.90 & 1.98 & 1.96 & 2.04 \\
KF & 1.51 & 1.63 & \textbf{1.54} & 1.67 & 1.81 & 1.83 & 1.88 & 1.89 & 1.93 & 1.96 & 2.03 \\
EKF & 1.66 & 1.65 & 1.60 & 1.78 & 1.88 & 1.86 & 1.90 & 1.91 & 1.96 & 1.99 & 2.04 \\
PF & \textbf{1.44} & 1.64 & 1.64 & 1.78 & 1.97 & 1.84 & 1.92 & 1.94 & 1.97 & 2.00 & 2.06 \\
DNN$^\dagger$ & 1.93 & \textbf{1.53} & 1.56 & \textbf{1.40} & \textbf{1.68} & \textbf{1.64} & \textbf{1.40} & \textbf{1.60} & \textbf{1.61} & \textbf{1.62} & \textbf{1.78} \\
\bottomrule
\end{tabular*}
\end{table}

\begin{table}[H]
\centering
\caption{Wall-clock runtime (s) of the cluster representative and post-clustering estimators across all evaluated values of $k$. For Rep, runtime includes the full clustering step; for all other estimators, it covers only the estimation step. Bold denotes the best (lowest) value per column; Rep is excluded from this comparison because its runtime includes the full clustering step.}
\label{tab:est_rt}
\renewcommand{\arraystretch}{1.15}
\small
\begin{tabular*}{\textwidth}{@{\extracolsep{\fill}}lrrrrrrrrrrr}
\toprule
\textbf{Estimator} & \textbf{10} & \textbf{50} & \textbf{100} & \textbf{150} & \textbf{200} & \textbf{250} & \textbf{300} & \textbf{350} & \textbf{400} & \textbf{450} & \textbf{500} \\
\midrule
\multicolumn{12}{l}{\textit{SDSGC (medoid)}} \\
Rep & 0.99 & 11.00 & 5.30 & 6.60 & 7.80 & 9.50 & 11.50 & 13.50 & 13.30 & 14.00 & 17.60 \\
KF & 0.18 & 0.59 & \textbf{0.87} & 1.29 & 1.51 & \textbf{1.53} & 1.76 & 1.91 & 2.10 & 2.09 & \textbf{2.36} \\
EKF & 0.17 & 0.58 & 0.91 & \textbf{1.16} & 1.54 & 1.55 & 1.80 & \textbf{1.85} & 2.04 & 2.14 & 2.46 \\
PF & \textbf{0.16} & \textbf{0.55} & 0.88 & 1.21 & \textbf{1.45} & \textbf{1.53} & \textbf{1.73} & \textbf{1.85} & \textbf{1.95} & \textbf{2.08} & 2.64 \\
DNN$^\dagger$ & 7.76 & 5.58 & 5.73 & 6.10 & 6.35 & 6.02 & 6.17 & 6.34 & 6.39 & 6.43 & 6.95 \\
\midrule
\multicolumn{12}{l}{\textit{UK-medoids (mountain)}} \\
Rep & 0.16 & 0.76 & 2.64 & 5.08 & 6.39 & 7.42 & 9.11 & 9.97 & 11.09 & 12.41 & 13.56 \\
KF & 0.19 & 0.63 & 1.01 & 1.42 & 1.61 & 1.80 & 1.95 & 2.33 & 2.22 & 2.46 & 2.56 \\
EKF & 0.19 & 0.64 & 1.00 & 1.51 & 1.84 & 1.80 & 2.03 & 2.27 & 2.43 & 2.47 & 2.76 \\
PF & 0.18 & 0.62 & 0.98 & 1.44 & 1.60 & 1.67 & 1.98 & 2.26 & 2.26 & 2.31 & 2.50 \\
DNN$^\dagger$ & 6.03 & 5.50 & 5.53 & 6.17 & 6.20 & 5.83 & 6.29 & 6.02 & 6.37 & 6.63 & 7.11 \\
\bottomrule
\end{tabular*}
\end{table}

\begin{table}[H]
\centering
\caption{Peak RAM (MB) of the cluster representative and post-clustering estimators across all evaluated values of $k$. For Rep, peak RAM includes the full clustering step; for all other estimators, it covers only the estimation step. SDSGC clustering RAM grows quadratically with $n$ due to the dense $n \times n$ affinity matrix. Bold denotes the best (lowest) value per column; Rep is excluded from this comparison because its peak RAM includes the full clustering step.}
\label{tab:est_ram}
\renewcommand{\arraystretch}{1.15}
\small
\begin{tabular*}{\textwidth}{@{\extracolsep{\fill}}lrrrrrrrrrrr}
\toprule
\textbf{Estimator} & \textbf{10} & \textbf{50} & \textbf{100} & \textbf{150} & \textbf{200} & \textbf{250} & \textbf{300} & \textbf{350} & \textbf{400} & \textbf{450} & \textbf{500} \\
\midrule
\multicolumn{12}{l}{\textit{SDSGC (medoid)}} \\
Rep & 21.6 & 281.5 & 599.0 & 921.0 & 1{,}205.0 & 1{,}449.0 & 1{,}662.0 & 1{,}853.0 & 2{,}030.0 & 2{,}191.0 & 2{,}340.0 \\
KF & 1.3 & \textbf{3.8} & \textbf{5.8} & \textbf{7.3} & \textbf{8.3} & 8.6 & \textbf{8.2} & 8.6 & 8.8 & 9.0 & 9.3 \\
EKF & 1.3 & \textbf{3.8} & \textbf{5.8} & \textbf{7.3} & \textbf{8.3} & 8.6 & \textbf{8.2} & \textbf{8.5} & \textbf{8.7} & \textbf{8.9} & \textbf{9.2} \\
PF & \textbf{1.2} & \textbf{3.8} & \textbf{5.8} & \textbf{7.3} & \textbf{8.3} & \textbf{8.5} & \textbf{8.2} & \textbf{8.5} & \textbf{8.7} & \textbf{8.9} & \textbf{9.2} \\
DNN$^\dagger$ & 65.8 & 65.9 & 65.8 & 66.2 & 66.5 & 66.6 & 66.9 & 67.0 & 67.2 & 67.3 & 67.5 \\
\midrule
\multicolumn{12}{l}{\textit{UK-medoids (mountain)}} \\
Rep & 6.3 & 6.5 & 7.0 & 7.5 & 8.0 & 8.5 & 9.0 & 9.5 & 10.0 & 10.5 & 11.0 \\
KF & 1.4 & 4.4 & 6.8 & 8.4 & 8.8 & 9.2 & 9.6 & 9.0 & 9.3 & 9.5 & 9.8 \\
EKF & 1.4 & 4.4 & 6.8 & 8.4 & 8.9 & 9.1 & 9.5 & 9.0 & 9.3 & 9.5 & 9.8 \\
PF & 1.4 & 4.4 & 6.8 & 8.4 & 8.9 & 9.1 & 9.5 & 9.0 & 9.3 & 9.5 & 9.8 \\
DNN$^\dagger$ & 65.8 & 65.9 & 66.2 & 66.5 & 66.9 & 67.1 & 67.4 & 67.6 & 67.8 & 67.9 & 68.0 \\
\bottomrule
\end{tabular*}
\end{table}

\end{document}